\documentclass[a4paper,fleqn]{cas-dc}

\usepackage[numbers]{natbib}
\usepackage{float}
\usepackage{graphicx}
\usepackage{amsmath,amssymb,amsfonts}
\usepackage{hyperref}
\usepackage[ruled,vlined]{algorithm2e}

\usepackage{multirow}
\usepackage{tabularx, booktabs}

\usepackage{subfig}
\usepackage{mathtools}
\usepackage{arydshln}
\usepackage{xcolor}
\usepackage{mwe}

\usepackage{xcolor}

\definecolor{red}{rgb}{1.00,0.00,0.00}
\definecolor{blue}{rgb}{0.00,0.00,1.00}

\def\tsc#1{\csdef{#1}{\textsc{\lowercase{#1}}\xspace}}
\tsc{WGM}
\tsc{QE}
\tsc{EP}
\tsc{PMS}
\tsc{BEC}
\tsc{DE}
\begin{document}

\let\WriteBookmarks\relax
\def\floatpagepagefraction{1}
\def\textpagefraction{.001}

\shorttitle{Lifelong Ensemble Learning for Few-Shot Object Recognition}
\shortauthors{H. Kasaei et~al.}

\title [mode = title]{Lifelong Ensemble Learning based on Multiple Representations for Few-Shot Object Recognition}      

% \tnotemark[1]

\author[1]{Hamidreza Kasaei}%[orcid=0000-0002-5418-6352]
\cormark[1]
\fnmark[1]
\ead{hamidreza.kasaei@rug.nl}
\ead[url]{https://www.ai.rug.nl/irl-lab/}

\address[1]{Department of Artificial Intelligence, Bernoulli Institute, Faculty of Science and Engineering, University of Groningen, The Netherlands.}

\author[2]{Songsong Xiong}
% \cormark[1]
\fnmark[1]
%\ead{s.xiong@rug.nl}

\address[2]{Department of Artificial Intelligence, Bernoulli Institute, Faculty of Science and Engineering, University of Groningen, The Netherlands.}

\fntext[1]{Equal contributions.}

\begin{abstract}
~{Service robots are increasingly integrating into our daily lives to help us with various tasks.} In such environments, robots frequently face new objects while working in the environment and need to learn them in an open-ended fashion. Furthermore, such robots must be able to recognize a wide range of object categories. In this paper, we present a lifelong ensemble learning approach based on multiple representations to address the few-shot object recognition problem. In particular, we form ensemble methods based on deep representations and handcrafted 3D shape descriptors. To facilitate lifelong learning, each approach is equipped with a memory unit for storing and retrieving object information instantly. The proposed model is suitable for open-ended learning scenarios where the number of 3D object categories is not fixed and can grow over time. We have performed extensive sets of experiments to assess the performance of the proposed approach in offline, and open-ended scenarios. For evaluation purposes, in addition to real object datasets, we generate a large synthetic household objects dataset consisting of $27000$ views of $90$ objects. Experimental results demonstrate the effectiveness of the proposed method on online few-shot 3D object recognition tasks, as well as its superior performance over the state-of-the-art open-ended learning approaches. Furthermore, our results show that while ensemble learning is modestly beneficial in offline settings, it is significantly beneficial in lifelong few-shot learning situations. Additionally, we demonstrated the effectiveness of our approach in both simulated and real-robot settings, where the robot rapidly learned new categories from limited examples. A video of our experiments is available online at: \href{https://youtu.be/nxVrQCuYGdI}{\texttt{\small{https://youtu.be/nxVrQCuYGdI}}} 
\end{abstract}
\iffalse
\begin{graphicalabstract}
\includegraphics{figs/grabs.pdf}
\end{graphicalabstract}
\fi

\iffalse
\begin{highlights}
\item Research highlights item 1
\item Research highlights item 2
\item Research highlights item 3
\end{highlights}
\fi

\maketitle

%%%%%%%%%%%%%%%%%%%%%%%%%%%%%%%%%%%%%%%%%%%%%%%%%%%%%%%%%%%%%%%%%%%%%%%%%%
\section{Introduction}

Nowadays many countries are facing labor shortages due to the aging of the population and the COVID-19 pandemic. Therefore, an ever-increasing amount of attention has been focused on service robots to handle such shortages. Intelligent robots require a variety of functionalities, such as perception, manipulation, and navigation, to safely interact with human users and environments~{\cite{INTR09}}. Among these functionalities, we believe that three-dimensional (3D) object perception plays a pivotal role~\cite{INTR_manzoor20213d}. This is because robots are often required to accurately identify the objects present in their surroundings and determine their locations to assist users in various tasks.

\begin{figure}[!t]
\vspace{2mm}
\centerline{\includegraphics[width=\linewidth]{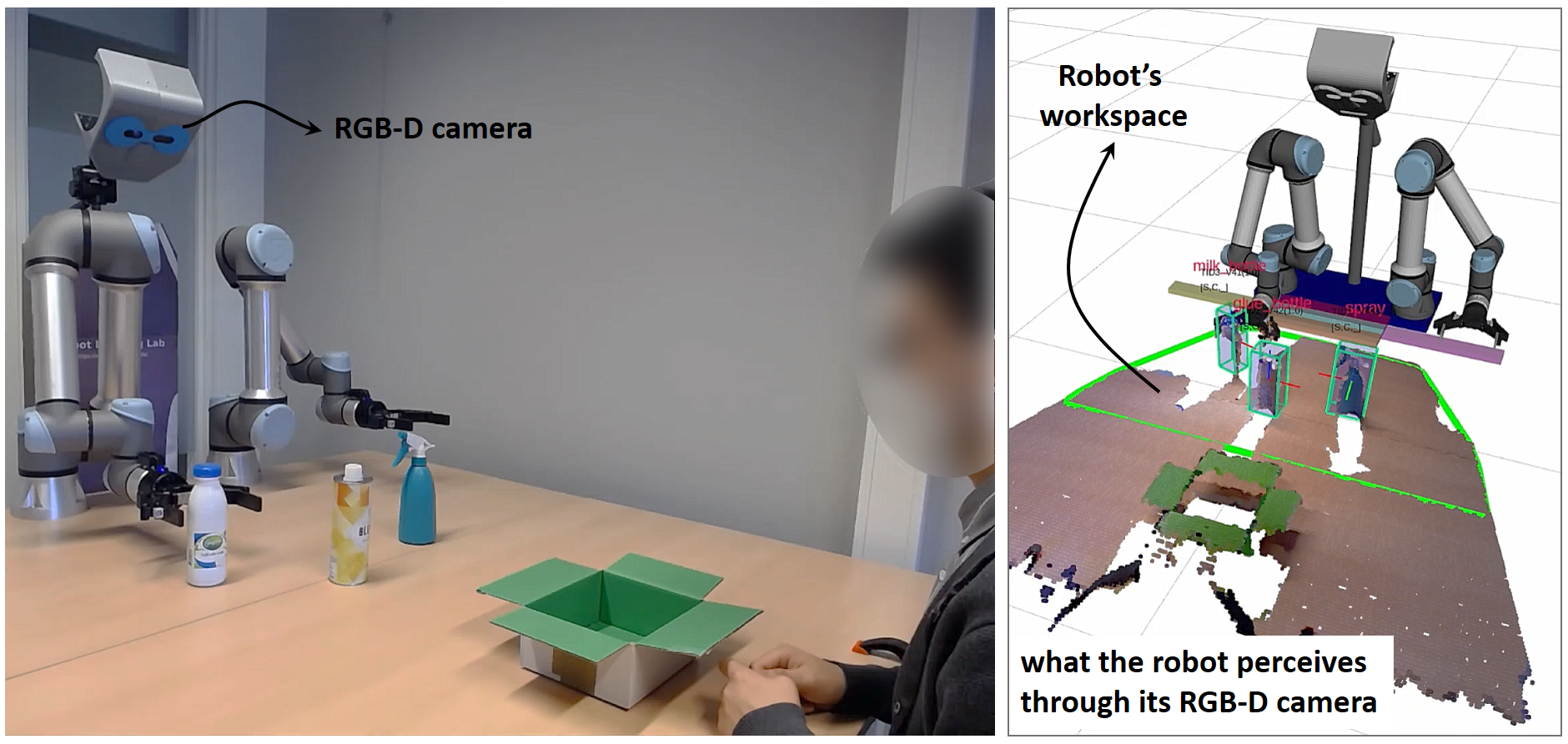}}
\caption{An example of robot-assisted packaging scenario: (\textit{left}) the dual-arm robot perceives the environment through an RGB-D camera, and then plans a collision-free trajectory to grasp the target object, and delivers it to the user; (\textit{right}) In order to successfully ~{hand over} a target object to a user, the robot should have a clear understanding of its current configuration (i.e., joint poses) and recognize all objects that are not part of the robot's body.}
\label{robot_assisted_packaging}
\end{figure}

As an example, consider a robot-assisted packaging scenario, where a robot is working alongside a human user (Fig.~\ref{robot_assisted_packaging}). In such scenarios, the user may ask the robot to ~{hand over} a specific object. To accomplish this task, the robot should be able to represent the current state of the environment in terms of objects' pose and category label, and then plan a collision-free trajectory to grasp the target object, pick it up, and deliver it to the user. In such collaborative settings, the robot often faces new objects over time and it is not feasible to assume that one can pre-program all object categories for the robot in advance. Furthermore, the human co-worker expects that the robot can easily adapt to different tasks by learning a new set of object categories based on a few instructions.

\begin{figure*}[!t]
\centerline{\includegraphics[width=\textwidth,trim=0 0 0 0,clip]{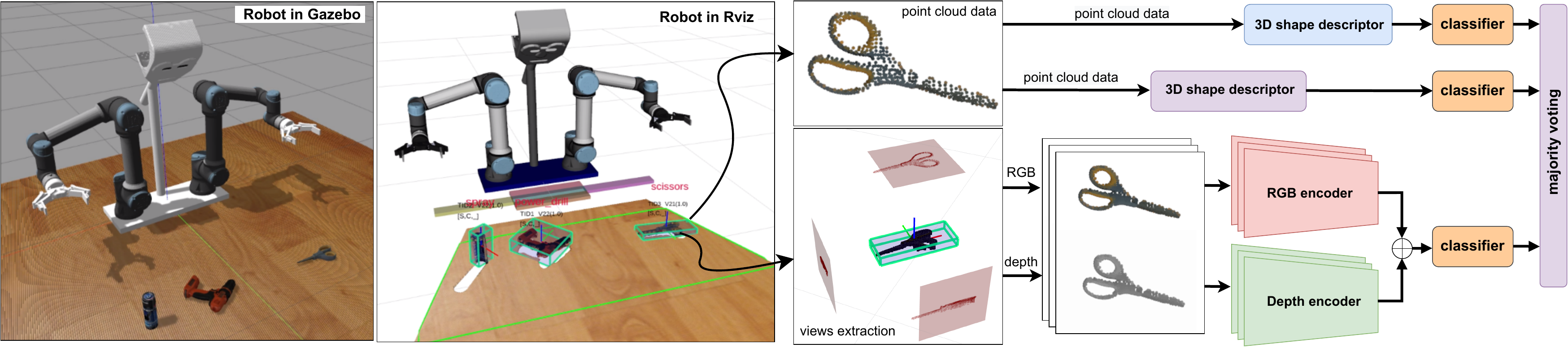}}
\caption{Overview of the ensemble learning process for 3D object recognition: In this example, the robot needs to detect and recognize a set of tools (spray, drill, and scissors. (\textit{left}) Our experimental setup in the Gazebo environment. (\textit{center}) We visualize the working area of the robot (green rectangle), ~{the bounding box of objects (green cube)}, and the recognition results (red text) in Rviz environment. (\textit{right}) The proposed ensemble learning method is constructed based on handcrafted shape descriptors and deep representations. The final classification result is obtained through a majority voting process.}
\label{ensemble_overview}
\end{figure*}

In such lifelong robot learning scenarios, object representation is crucial because the output of this module is used for both learning and recognition~{\cite{INTR10}}. Most of the object representation approaches encode either geometric features~\cite{INTR06} or textural features~\cite{INTR07} of an object. However, there are a few works that encode both geometric and textural features simultaneously~\cite{INTR08}. Each of these approaches has its own cons and pros. Consequently, to benefit from all of these approaches, an ensemble learning method can be developed, where multiple object recognition methods based on individual representation are trained and their predictions are combined~{\cite{INTR11}\cite{INTR12}}.~{Ensemble learning not only reduces the rate of misclassifications but also enhances predictions compared to individual models. However,}

the cost of ensemble learning increases with the number of integrated approaches. Therefore, a trade-off between recognition accuracy and computation time should be considered, otherwise, it becomes computationally untenable for robotics applications.

In this paper, we present a robust lifelong ensemble learning method that enables robots to learn new object categories over time using few-shot training instances. ~{Particularly}, we consider multiple representations to encode different features of the objects. Instead of concatenating all features and ~{training} a single model, we train a model for each object representation. The category label of the target object is finally determined based on the majority vote of contributing models. Figure~\ref{ensemble_overview} shows an overview of our work. To assess the performance of the proposed approach, we ~{perform} extensive sets of experiments concerning recognition accuracy, scalability, robustness to the Gaussian noise and downsampling.  ~{Experimental results show that our approach outperformed the selected state-of-the-art methods in terms of classification accuracy and scalability.} Additionally, the proposed approach demonstrated better robustness compared to other approaches under low and mid-level noise or downsampling conditions. In summary, our key contributions are threefold:

\begin{itemize}
    \item We proposed a lifelong ensemble learning method based on multiple deep and handcrafted representations for few-shot object recognition. 
    \item We extensively evaluated the performance of the proposed approach in offline, open-ended, and robotics scenarios. Furthermore, we assessed the robustness of our method against various ~{levels} of Gaussian noise and downsampling. Experimental results showed that while ensemble learning of different representations is modestly beneficial in offline situations, it is significantly beneficial in online learning situations.
     \item We generated a large synthetic household objects dataset that consists of $27000$ views of $90$ objects. To the benefit of research communities, we make the dataset publicly available online. 
    
\end{itemize}

%%%%%%%%%%%%%%%%%%%%%%%%%%%%%%%%%%%%%%%%%%%%%%%%%%%%%%%%%%%%%%%%%%%%%%
\section {Related Work}
\label{sec:related_work}
In this section, we provide a brief overview of recent efforts in 3D object recognition, active learning, and ensemble learning. However, it should be noted that a comprehensive review of these topics is beyond the scope of this work.
 
\subsection{3D Object Representation}

The development of robust 3D object representation has been a subject of research in computer vision, machine learning, and robotics communities. In general, object descriptors can be categorized in two main categories: \textit{local} and \textit{global} descriptors. Local descriptors encode a small area of an object around a specific key point, while global descriptors describe the entire 3D object~\cite{Relate01}. In general, global descriptors are more efficient in terms of computation time and memory usage than local descriptors, which makes them more suitable for real-time robotic applications. When it comes to robustness to occlusion and clutter, local descriptors perform better than global descriptors~\cite{Relate05}. 

Classical studies focused on handcrafted 3D object descriptors to capture the geometric essence of objects and generate a compact and uniform representation for a given 3D object~\cite{Relate02}. For instance, Wohlkinger and Vincze~\cite{Relate03} proposed Ensemble of Shape Function (ESF) to encode geometrical characteristics of an object using an ensemble of ten 64-bin histograms of angles, point distances, and area shape functions. In \cite{Relate04}, Viewpoint Feature Histogram (VFH) was presented for 3D object classification that encodes geometrical information and viewpoint. Kasaei et al.,~\cite{Relate05} introduced an object descriptor called Global Orthographic Object Descriptor (GOOD) built to be robust, descriptive and efficient to compute and use. For a detailed review of various handcrafted object descriptors, we refer the reader to a comprehensive review by Carvalho and Wangenheim \cite{Relate02}.

In addition to handcrafted descriptors, there are several (Bayesian or deep) learning based approaches to encode different properties of objects. In the case of Bayesian learning based features, Kasaei et al.,~\cite{Relate06} extended Latent Dirichlet Allocation (LDA) and introduced Local-LDA, which is able to learn structural semantic features for each category independently. Ayoubi et al.,~\cite{Relate07} extended the local-LDA and introduced Local-HDP, a non-parametric hierarchical Bayesian approach for 3D object categorization. In particular, the advantages of handcrafted features and the learning-based structural semantic features have been considered in such LDA-based approaches. 

Deep learning based object representation methods can learn a descriptive representation of the objects in either supervised~\cite{Relate26}, or unsupervised~\cite{Relate09} fashions.~{These approaches can be classified into three different categories based on their input: \textit{volume-based}, \textit{point-based}, and \textit{view-based} approaches. Voxel-based and point-based methods employ different representations for 3D objects—3D meshes and collections of unordered points, respectively. Both approaches commonly utilize 3D networks to tackle the challenge of 3D object recognition. While these 3D networks comprehensively capture the inherent spatial information within these objects, their practicality is hindered by the high computational costs involved~\cite{INTR_zhou2019multi}. Contrary to voxel-based and point-based techniques, view-based methods are the most effective in 3D object recognition~\cite{Relate10}. View-based methods render different 2D images of a 3D object by projecting the object’s points onto 2D planes~\cite{Relate11}. The obtained views are then used to train the network \cite{Relate12}. Compared to the previous two methods, view-based approaches are able to contribute to the enhancement of computational efficiency.  However, within the traditional deep training process, these methods often require a large number of data, and long training time to achieve satisfactory results.} To overcome these limitations, transfer learning approaches have been suggested~{~\cite{Relate13, Relate27}}.

In this paper, we consider both handcrafted descriptors and multi-view deep transfer learning-based representations in the context of lifelong ensemble learning for 3D object recognition. {Handcrafted descriptors often demonstrate resilience to spatial scale variations and rotations within certain limits~\cite{bento2022comparing} as they are manually designed by engineers to handle such cases. However, their efficacy diminishes when confronted with complicated scenes and partial occlusions. In contrast, deep learning-based representation autonomously learns features from data, enabling them to navigate through complex scenarios and manage partial occlusions effectively ~\cite{alzubaidi2021review}. Despite these advantages, deep models often require substantial data and may exhibit sensitivity to spatial scale variations. The proposed solution advocates for a combined approach, leveraging the robustness of handcrafted descriptors and the adaptability of deep representations. This integration enhances model performance by considering the complementary strengths of both methodologies.} 

{\subsection{Ensemble learning}}
{In the field of object recognition, ensemble learning has emerged as a promising approach to enhance recognition performance using base models. Various ensemble learning methodologies such as bagged ensembles~\cite{breiman2001random,buhlmann2002analyzing}, boosted ensembles~\cite{han2016incremental,liu2014facial,moghimi2016boosted,walach2016learning}, and stacked ensembles~\cite{mane2021handwritten,deng2012scalable} have been successfully employed to address object recognition tasks. These techniques enable the combination of multiple models to form an ensemble, which can be trained to produce more accurate and robust predictions than any of the individual models on their own. 
In recent years, deep learning has become the predominant approach for classification tasks, and its success has been largely attributed to the availability of large datasets, such as ImageNet. However, the performance of deep learning models is significantly limited in scenarios where the amount of available data is scarce, such as in few-shot object recognition. This is due to the fact that deep learning models require a vast amount of data to train effectively, and without sufficient data, their performance suffers. As a result, alternative approaches, such as ensemble learning, have been proposed to improve the recognition performance of deep learning models in few-shot scenarios. Ensemble convolutional neural networks (CNNs) have been proposed and increasingly applied in classification tasks~\cite{Relate23,dong2022deep,chen2019deep,he2020transferring,kumar2016ensemble,yang2021two}. Such studies demonstrate that by using ensemble learning techniques, we can improve the performance of object recognition, and outperform individual transfer learning models. To the best of our knowledge, ~{there is currently no research that investigates} the impact of ensemble learning based on CNNs and handcrafted representations on few-shot 3D object recognition. Unlike the previous approaches, we primarily focus on utilizing bagging ensemble with majority voting strategies for few-shot object classification. In particular, we investigate the effectiveness of ensemble learning based on CNNs and handcrafted representations in few-shot object recognition and provide insights into how the ensemble models can improve recognition performance in scenarios where data is limited.}

%=====================================
\subsection{Active Object Recognition}

The performance of the deep 3D object recognition approach is heavily dependent on the quantity and quality of training data. The process of annotating data is also labor-intensive and expensive. 
Catastrophic interference/forgetting is another important limitation of deep approaches~\cite{Relate14, Relate31}. Furthermore, the scarcity of data in some categories severely limits the application of deep learning methods.

Several active, few-shot, and ensemble learning approaches have been proposed in order to overcome these limitations. Most active learning (AL) methods aim to update the known model and achieve a certain recognition accuracy by seeking the minimal training example~\cite{Relate15,Relate16,Relate17,Relate18,Relate19}. In particular, a subset of training examples is first selected from the unlabeled pool of data, by using an acquisition function based on either uncertainty measures or density/geometric similarity measures in feature space. After that, the samples are labeled by an oracle.~{Lastly,  within the realm of incremental learning, many method~{~\cite{wu2019large,castro2018end,paul2022class,li2017learning}} assume that the distribution of data for existing classes remains static over time. However, this assumption may not align with the dynamic nature of real-world scenarios, where data distributions are susceptible to change.} Besides, the new information is incrementally incorporated into the model or re-trained from scratch without creating a catastrophic result~{~\cite{Relate29,Relate30}}. These approaches are incremental rather than lifelong since the number of categories is predefined in advance.~{Additionally, ~\cite{liu2021l3doc} and ~\cite{sun2021and} also tackle the challenges of lifelong learning by capturing and transferring task-relevant knowledge to adapt to new tasks. However, the performance of these ~{methods} in the current task is influenced by the degree of relevance to previous tasks. }

In the case of few-shot learning, the agent initially has access to a large training dataset, $D_{train}$, to learn a representation function, $f$. During the testing phase, the agent is given~{a new few dataset ($D_{few}$)}, consisting of samples from a new out-of-the-box distribution, to learn and recognize a set of new object categories. Note that $D_{train}$ has no known relationship to $D_{few}$ (i.e., cross-domain learning). Several studies have shown that the best approach to few-shot learning is to use high-quality feature ~{extractors} rather than complicated meta-learning (i.e., learning to learn) algorithms~\cite{Relate20,Relate21, Relate33}. While we agree with this point, we believe that one representation is insufficient to cover a variety of distributions and the performance can be improved with an ensemble of multiple (deep and handcrafted) representations.  ~{There are some works that address an ensemble of deep networks for few-shot classification \cite{Relate22,Relate23,Relate24}}. These approaches are computationally very expensive in both the training and testing phases, hence, it is challenging to use them in real-time robotic applications. One way to reduce overhead at the test time is to train another network to imitate the behavior of the ensemble (i.e., distillation of an ensemble into a single model~\cite{Relate25}). Applying distillation in lifelong learning scenarios is a very challenging task. Unlike these approaches, we proposed lifelong ensemble learning based on multiple handcrafted and deep representations to tackle real-time 3D object recognition tasks.  

%%%%%%%%%%%%%%%%%%%%%%%%%%%%%%%%%%%%%%%%%%%%%%%%%%%%%%%%%%%%%%%%%%%%%%%%%%
\section {Method}

{As shown in Fig.~\ref{ensemble_overview}, the robot used an RGB-D sensor (i.e., ASUS Xiton camera) to capture a point cloud of the environment. 
The robot then segments the tabletop objects and sends the point cloud of each object into the recognition pipeline.} Our approach receives the point cloud of an object and produces multiple handcrafted and deep representations for the given object. The obtained representations are then used to train different classifiers in a lifelong learning setting. As opposed to sampling and labeling the training data in advance, we propose to iteratively and adaptively determine which training instance should be labeled next. In particular, we follow an active learning scenario, in which a human user is involved in the learning loop to teach new categories or to provide feedback in an online manner. In our approach, all classifiers are equivalent, and a majority voting scheme is used as the aggregation function (each classifier votes for one hypothesis). {It should be noted that in this study, we will assume that an object has been segmented from the scene, and we will concentrate on 3D object recognition. Our approach works directly on 3D point clouds (i.e., containing RGB and Depth data) and does not need object point triangulation or surface meshing.} In the following subsections, we discuss our approach in detail. 

%=====================================

\subsection{Object Representation}

A point cloud of an object consists of a set of points, $\mathbf{p}_{i}: i \in\{1, \ldots, n\}$ where each point is represented by its 3D coordinate $[x,y,z]$ and RGB information $[r,g,b]$ ~{~\cite{Relate28}}. We intend to encode both textural and geometrical features of objects using handcrafted shape descriptors and deep transfer learning approaches, respectively. Among all possible handcrafted 3D shape descriptors, we consider GOOD~\cite{Relate05}, ESF~\cite{Relate03}, VFH~\cite{Relate04}, and GRSD~\cite{Relate08} descriptors to encode the geometrical properties of the object. All of these descriptors showed very good descriptiveness power, scalability, and efficiency (concerning both memory footprint and computation time) for real-time robotics applications~\cite{Relate05}.

{To capture the texture of the 3D object, we employ the OrthographicNet method. This approach involves generating three orthographic views of the object and subsequently passing these views through a pre-trained network~\cite{Result18}. The process entails the construction of a local reference frame for the object, followed by the generation of multiple RGB and Depth views from various perspectives. This methodology allows us to effectively encode the texture information of the 3D object, enhancing our ability to represent its visual characteristics in a comprehensive manner.} Particularly, the geometric center of the object is first computed by arithmetically averaging all points of the object. Then, we apply principal component analysis (PCA) on the point cloud of the object to find three eigenvectors, $\left[\mathbf{v}_{1}, \mathbf{v}_{2}, \mathbf{v}_{3}\right]$, sorted by eigenvalues, $\lambda_{1} \ge \lambda_{2} \ge \lambda_{3}$ of the object. Therefore, the $\mathbf{v}_{1}, \mathbf{v}_{2}, \mathbf{v}_{3}$ eigenvectors are considered as X, Y, Z axes of the object respectively. The object is subsequently transformed to be placed in the obtained reference frame. {To generate orthographic RGB-D images from a 3D object, we position three virtual RGB-D cameras strategically around the object. These cameras are oriented such that the Z-axis of each camera points toward the centroid of the object. Additionally, the Z-axis of each camera is perpendicular to the first two axes of the object and parallel to the third axis. Subsequently, a ray-tracing technique is employed to render both depth and RGB images of the object., regardless of how accurate/complete the point cloud of the object is. The obtained images are then passed into a pre-trained Convolutional Neural Network (CNN) to obtain a deep representation of the 3D object. Towards this goal, the topmost classification layer of the CNN model is omitted. Subsequently, both RGB and Depth images are fed into the CNN model to compute embeddings for each image. The obtained embeddings are then concatenated to construct a deep representation of the object, which is further employed to train a classifier for ensemble classification}~An illustrative example of this procedure is shown in Fig.~\ref{ensemble_overview}, where we generated orthographic views for the scissors.

\subsection {Object Category Learning and Recognition}
~{In numerous robotics applications, it is crucial for the robot to learn and recognize objects based on a few training data~{\cite{Method01}}}. The problem becomes even more complicated when the robot must learn new categories online, while still maintaining the recognition performance on previously learned categories. Consider an example where a robot faces a new object while working in the environment. In such situations, the robot should be able to learn the new category on-site by observing a few examples of the object, without accessing the old training data and re-training from scratch. This point can be addressed ~{through} lifelong learning, where, the number of categories increases over time based on the robot's observations and interactions with humans.  {We have tackled this problem by training a set of {\textit{independent}} Instance-Based Learning (IBL) models based on multiple representations. In our approach, the model makes predictions based on specific instances from the training dataset rather than relying on generalized patterns. In IBL, the model memorizes the training instances and uses them directly for predictions.} It is worth mentioning that aggregating different features and classifying the obtained representation with a unified model is not beneficial. Furthermore, combining feature vectors of different lengths is a non-trivial task as standard pooling functions such as average pooling cannot be used directly. Although we can consider dimensionality reduction techniques to address this issue, applying such techniques in each of the representation are computationally expensive and are not suitable for real-time robotic applications. Another alternative is to concatenate various feature vectors to form a single representation. Such an approach is also undesirable due to the resulting in a high dimensional feature vector, which can lead to the curse of dimensionality and poor performance. Therefore, we decided to use an ensemble method based on individual classifiers for different feature vectors, and then combine the decisions with a majority voting scheme.

In particular, we form a category by ~{a} set of known instances, $\textbf{C} \longleftarrow \{\textbf{o}_1, \dots,\textbf{o}_n\}$, where $\textbf{o}_i$ is the representation of an object view. The robot can learn a new category or update the model of an existing category by interacting with a user. {In particular, when the dissimilarity between the currently known categories and a test object surpasses a predefined threshold, the robot infers that the test object could be a novel object. Specifically, it deduces that the test object could be a novel instance of an already known category or an entirely new category. In response to this inference, the user is prompted to introduce the object to the robot through either teaching or correcting actions.}

Similar to \cite{Relate19}, we follow a ~{user-centered} labeling strategy where the \textit{teach} and \textit{correct} actions by the user trigger the robot to store a new instance of a specific category in the memory. Note that the robot can learn object categories in a ~{self-supervised manner}.~{For ~{instance}, when the dissimilarity between the current object being recognized and the already learned $m$ targets is above a certain threshold, the robot infers the currently recognizing objects as a new object category, namely ``category $m+1$''.} The problem is that such a category label is not meaningful for a human user, and the user cannot naturally instruct the robot to perform a task, such as ``\textit{bring me a cup of tea, please!}''.  {In online scenarios, particularly during open-ended evaluations or robot experiments, if the robot misclassifies an object -- meaning a known object is incorrectly treated as unknown, or an unknown object is erroneously identified as already known -- an (simulated) oracle provides corrective feedback. This feedback serves as a mechanism for error correction, allowing the robot to update the object category model. This iterative process of receiving corrective feedback and updating the model enables the robot to continually refine its knowledge and adapt to new or evolving object categories in real-time scenarios. Although the training phase is very fast, the robot should avoid storing redundant instances as it increases memory usage and slows down the recognition process.}

For recognition purposes, we assess the performance of various classifiers, including: k-Nearest Neighbors (k-NN), Decision Tree (DT), Quadratic Discriminant Analysis (QDA), Support Vector Machine (SVM), Gaussian Naive Bayes (GNB), and Multi-layer Perceptron (MLP) using ~{K-fold cross validation} and Restaurant object dataset~\cite{Result07}. Based on the obtained results, we use the k-NN classifier~{\cite{Relate32}} (see section~\ref{various_classifiers}). Therefore, each IBL approach can be considered as a combination of a category model and a dissimilarity measure. Since objects are represented as global features (histograms), the dissimilarity between two objects can be computed by various distance functions. In this paper, we investigate the effect of $10$ different distance functions on object recognition performance. A majority voting scheme is then used to combine the predictions of ensemble members to form the final prediction. In the event of a tie vote, the class that has the minimum distance to the query object is selected as the winner (see Fig.~\ref{ensemble_overview}).

%%%%%%%%%%%%%%%%%%%%%%%%%%%%%%%%%%%%%%%%%%%%%%%%%%%%%%%%%%%%%%%%%%
\section{Results and Discussions}

We performed several experiments to assess the performance of the proposed approach. In this section, we first present a detailed description of the datasets used for evaluation purposes, ~{followed by analyses of the various rounds of experiments in an offline, open-ended setting}. Additionally, we demonstrate the performance of our approach in real time through a series of robotic demonstrations. 

\begin{figure}[!t]
\includegraphics[width=\linewidth]{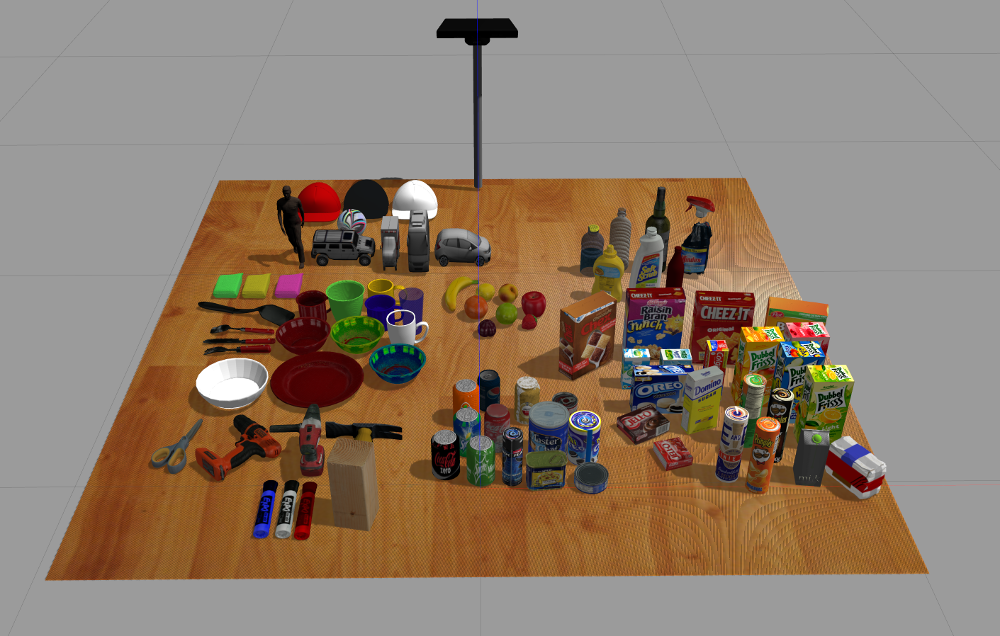}
\includegraphics[width=\linewidth, trim= 0mm 0mm 0mm -1mm, clip = true ]{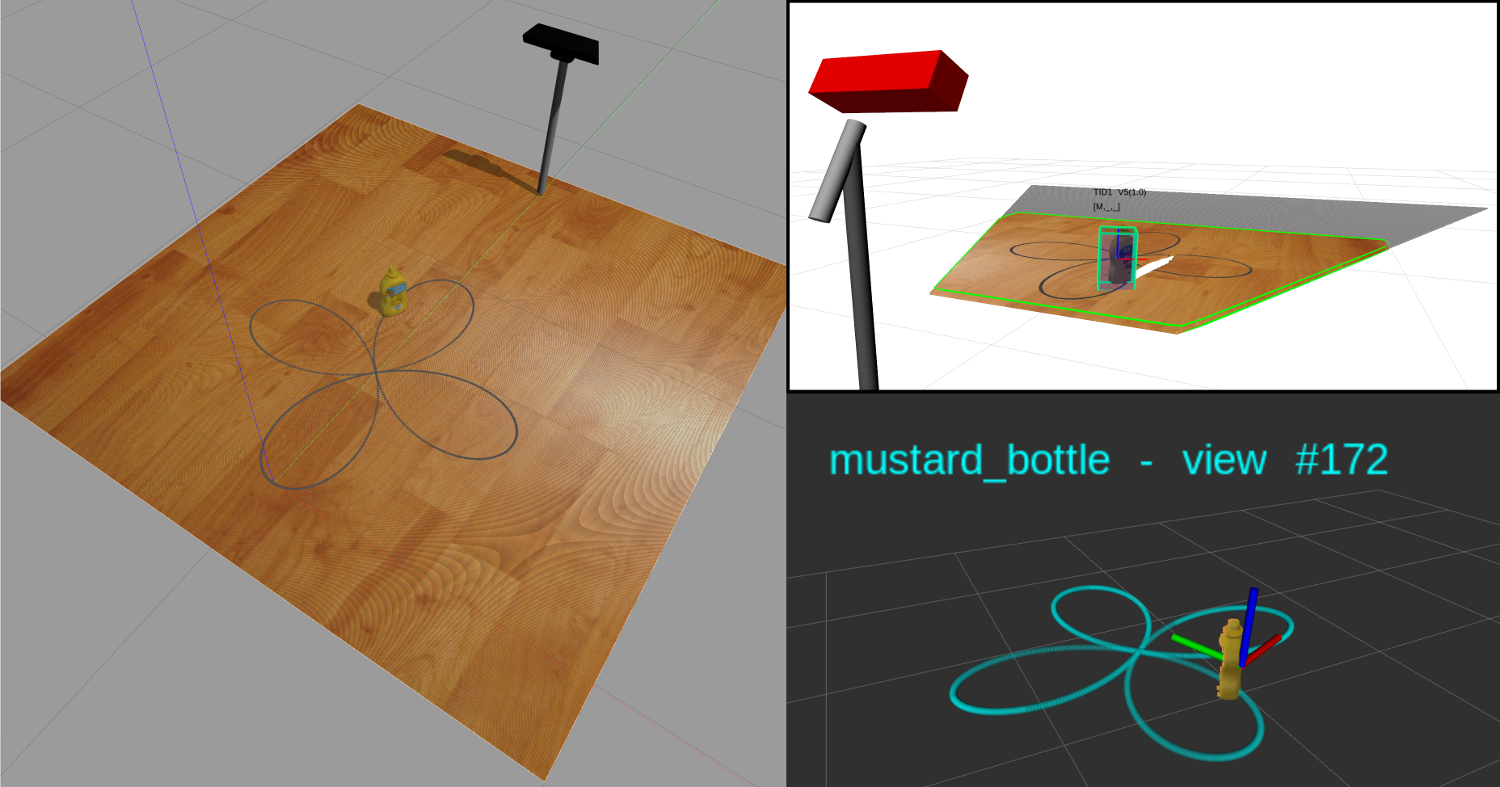}
\caption{Synthetic point cloud object dataset: (\textit{top-row}) synthetic household objects in Gazebo environment; (\textit{lower-row}) our simulation environment consisting of a Kinect camera and a table; To capture partial views of the target object, we rotate and move the object in a rose trajectory in front of the camera. %{(find an acronym for the dataset)}
}
\label{dataset}
\end{figure}
\subsection{Datasets}

\begin{figure*}[!t]
 \begin{tabular}{ccc}
        \hspace{-0.35cm} \includegraphics[width=0.35\linewidth]{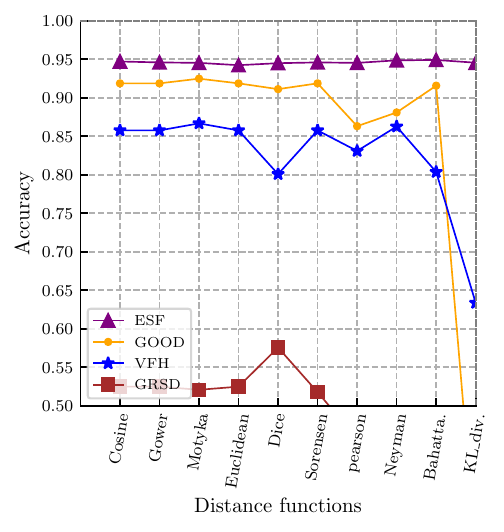} & \hspace{-0.8cm}
        \includegraphics[width=0.33\linewidth, trim= 0mm 0mm 0mm 0mm, clip = true]{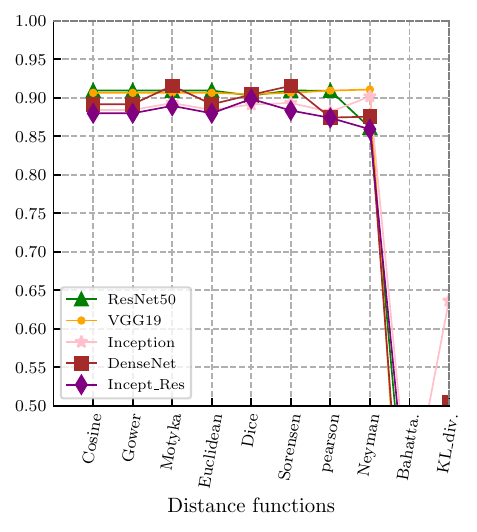} &\hspace{-0.8cm}
        \includegraphics[width=0.358\linewidth, trim= 0mm 0mm 0mm 0mm, clip = true]{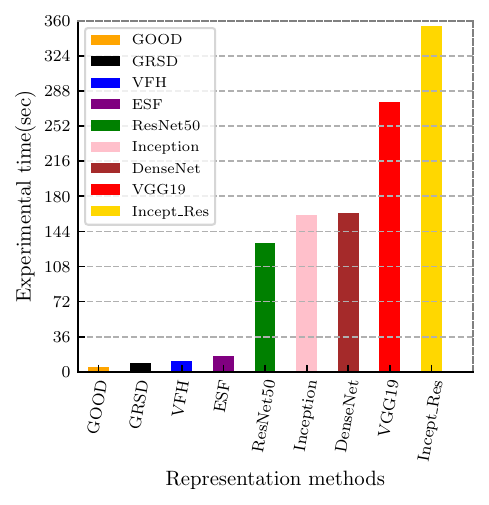}
    \end{tabular}
\caption{Performance of all handcrafted and deep learning descriptors over different distance functions on the restaurant object dataset~{in offline scenario}: (\textit{left}) recognition accuracy of handcrafted descriptors; (\textit{center}) recognition accuracy of deep learning based representations; (\textit{right}) {The experimental time for each approach signifies the duration needed to execute a 10-fold cross-validation experiment.}} 
\label{acc_different_functions}
\end{figure*}
As shown in Fig.~\ref{dataset} (\textit{top-row}), we developed a simulation environment in Gazebo to record a large synthetic household object dataset. Towards this goal, we considered $90$ simulated household objects, imported from different resources (e.g., the YCB dataset~\cite{Relate15}, Gazebo repository, etc.). It should be noted that this is a very challenging dataset for object recognition tasks since we include both basic-level (i.e., objects that are not similar to each other such as \textit{apple} vs. \textit{book}) and fine-grained (the object that ~{is} very similar together \textit{spoon} vs. \textit{fork}) object categories. Furthermore, there are several objects with the same geometry, but different textures, and vice versa. {To capture partially visible point clouds of an object at various scales, we employ a method where the object is moved along a rose trajectory in front of the camera. We record $300$ views for each object, as illustrated in Fig.~\ref{dataset} (lower row). This process enables the acquisition of diverse views of the object in different scales and orientations.} The obtained $27000$ partial views are then organized into $90$ object categories\footnote{\href{https://tinyurl.com/2at844tt}{GitHub repository: https://tinyurl.com/2at844tt}}.  

The synthetic household object dataset is used not only to select the optimal parameters for the proposed method, but also to assess the performance of other state-of-the-art approaches and comparison purposes. Furthermore, we used two real object datasets. In particular, we used one large-scale and one small-scale real object ~{dataset}, including the Washington RGB-D object dataset~\cite{Result02} and the Restaurant object dataset~\cite{Result07}. The former one contains $250000$ views from $300$ common household objects, organized into $51$ categories, whereas the ~{latter} has only $350$ views of objects that are categorized into $10$ categories with significant intra-class variations. This makes it a good choice for conducting extensive set of offline experiments.

\begin{figure*}[!t]
    % \centering
    \begin{tabular}{ccc}
        \hspace{-0.35cm} \includegraphics[width=0.35\linewidth]{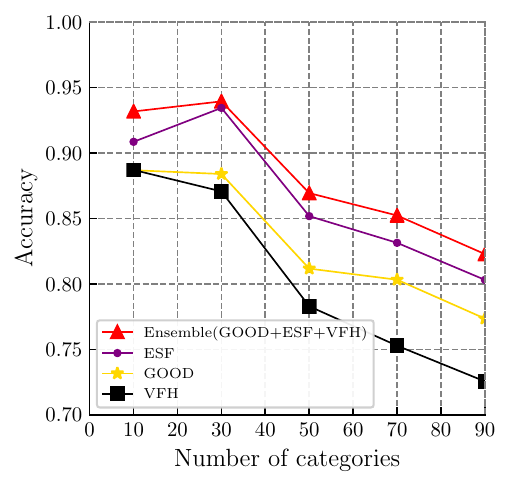} & \hspace{-0.6cm}
        \includegraphics[width=0.33\linewidth, trim= 0mm 0mm 0mm 0mm, clip =true]{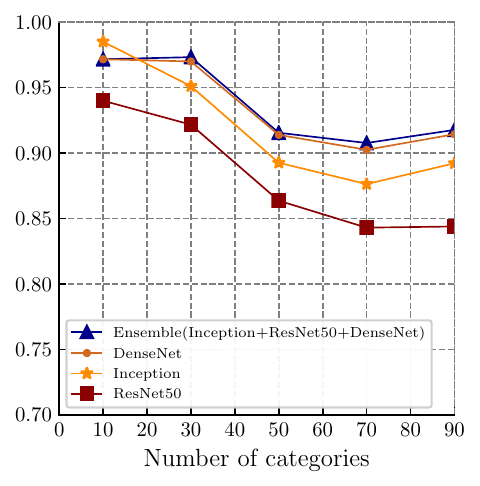}& \hspace{-0.6cm}
        \includegraphics[width=0.33\linewidth, trim= 0mm 0mm 0mm 0mm, clip =true]{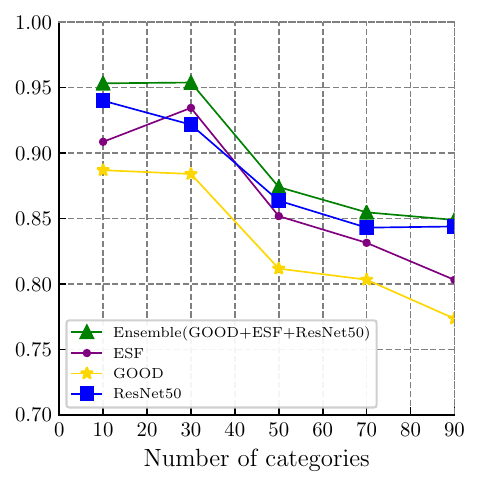} 
    \end{tabular}
    \caption{Scalability of the selected approaches has been measured as a function of accuracy versus number of categories~{on the synthethic household object dataset in offline scenario}: (\textit{left}) only handcrafted representations; (\textit{center}) only deep representations (\textit{right}) mixture of handcrafted and deep representations.}
    \vspace{-3mm}
    \label{scalability_of_ensambles_plot}
\end{figure*}

\begin{figure*}[!t]
    % \centering
    \begin{tabular}{ccc}
        \hspace{-0.35cm} \includegraphics[width=0.35\linewidth]{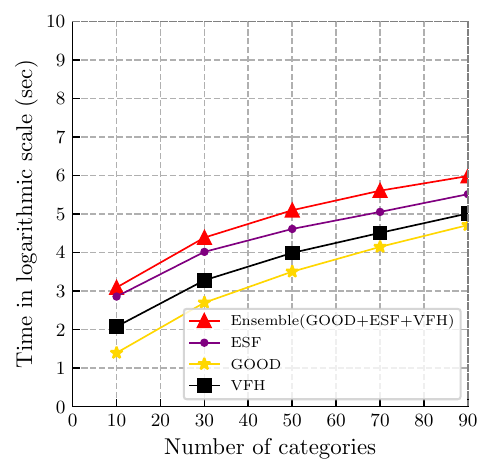} & \hspace{-0.6cm}
        \includegraphics[width=0.33\linewidth, trim= 0mm 0mm 0mm 0mm, clip = true]{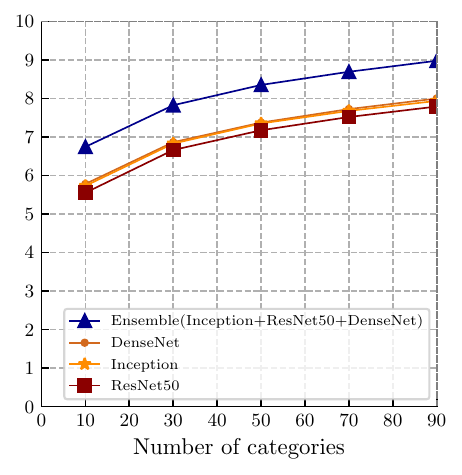} 
        & \hspace{-0.6cm}
        \includegraphics[width=0.33\linewidth, trim= 0mm 0mm 0mm 0mm, clip = true]{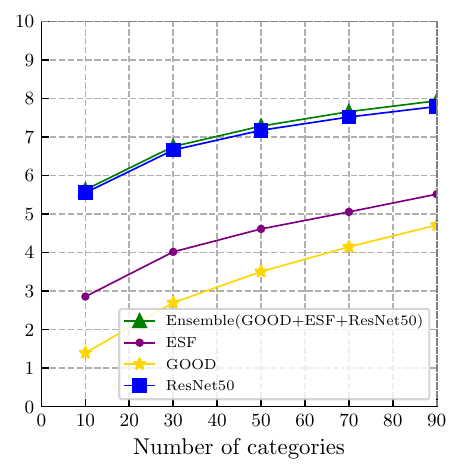}

    \end{tabular}
    \caption{Computation time of the selected approaches with respect to varying numbers of categories~{on the synthetic household object dataset in offline scenario}: (\textit{left}) only handcrafted representations; (\textit{center}) only deep representations (\textit{right}) mixture of handcrafted and deep representations.}
    \label{Time_Hc_Ensemble}
\end{figure*}
%==========================================
\subsection{Offline Evaluation}\label{AA}

Several experiments were carried out to evaluate the performance of the proposed approach concerning descriptiveness, scalability, robustness to Gaussian noise and downsampling, and computation efficiency.~{In this round of offline experiments, we used a $10$-fold ~{cross-validation} algorithm~\cite{Result23}. In each iteration, one fold is used as test data, and the remaining nine folds are used for training. } The process of cross-validation is then repeated $10$ times, and each fold is only used once as the test data.

\begin{table}
\newcolumntype{?}{!{\vrule width 0.5pt}}
\setlength\arrayrulewidth{0.5pt}
    \caption{Summary of the performance of various classifiers~{on the restaurant object dataset in offline scenario}.}
    \begin{center}
        \resizebox{\linewidth}{!}{
        \begin{tabular}{|c|c|c|c|c|}
        \hline
        \multirow{2}{*}{\textbf{Classifiers}}&\multicolumn{4}{c|}{\textbf{object representation methods}} \\
        \cline{2-5} 
         & \textbf{{GOOD}}& \textbf{{ESF}}& \textbf{{ResNet50}}& \textbf{{DenseNet}}\\
        \hline
        GNB&  0.8571 & 0.9217 & 0.8893 & 0.9129 \\
        \hline
        DT& 0.8872& 0.9066& 0.9021& 0.7975 \\
         \hline
        k-NN & \textbf{0.9248} & \textbf{0.9453} & \textbf{0.9094} & \textbf{0.9155} \\
        \hline
     QDA& 0.6574& 0.4024& 0.2956& 0.2921 \\
      \hline
        MLP& 0.7902& 0.7043& 0.2897& 0.5471 \\
      \hline
        SVM& 0.7553& 0.5678& 0.1868& 0.3244 \\
        
        \hline
        \end{tabular}}
    \label{classifiers}
    \end{center}
\end{table}
%=====================================================
 \subsubsection{Evaluation of different classifiers}
 \label{various_classifiers}
We performed a set of  ~{K-fold cross-validation experiments} to assess the performance of various classifiers on the Restaurant object dataset, and we chose the best one to serve as the base learner's classifier. In particular, we considered k-Nearest Neighbors (k-NN), Quadratic Discriminant Analysis (QDA), Support Vector Machine (SVM), Gaussian Naive Bayes (GNB), DecisionTree (DT), and Multi-layer Perceptron (MLP). In this round of experiment, we set k for k-NN to 3, and used the default parameters for other classifiers. Results are summarized in Table~\ref{classifiers}. 

By comparing all the obtained results, it is visible that k-NN achieved the best results using all representations. We ~{hypothesize} that object categories can be divided more linearly in high-dimensional spaces, therefore, classifiers like k-NN and naive Bayes offer superior classification and generalization performance.

%=====================================================
\subsubsection{Evaluation of individual representations}
We evaluated the recognition accuracy of several handcrafted descriptors and deep learning representations on the Restaurant Object Dataset to find out a set of good IBL methods to construct our ensemble approach. In the case of handcrafted approaches, we considered ESF~\cite{Relate03}, VFH~\cite{Relate04}, GOOD~\cite{Relate05}, and GRSD~\cite{Relate08} descriptors, and for deep learning representations, we evaluated Resnet50~\cite{Result01}, Inception~\cite{Result13}, DenseNet~\cite{Result10}, Inception-ResNet (Incept-Res)~\cite{Result12} and VGG19~\cite{Result11}. Each of these descriptors has its own parameters that should be optimized to provide a good balance among recognition accuracy, memory usage, and computation time. Furthermore, as mentioned earlier, the distance functions ~{play} a significant role in IBL methods. Therefore, we evaluated the impact of $10$ different distance functions including \textit{Cosine}, \textit{Gower}, \textit{Motyka}, \textit{Euclidean}, \textit{Dice}, \textit{Sorensen}, \textit{Pearson}, \textit{Neyman}, \textit{Bhattachayya} (Bahatta.), \textit{KL-divergance} (KL\_div.) on recognition accuracy. We use the Average Class Accuracy (ACA) as a means of deciding which performed better. In particular, ACA$= \frac{1}{K} \sum_{i=1}^K{acc_i}$, where $K$ is the number of classes in the dataset, and the accuracy of each class is calculated as $\frac{\#\operatorname{true~predictions}}{\#\operatorname{test samples}}$. We performed a grid search to determine the optimal hyper-parameters for each approach.

~{The GOOD descriptor has a parameter called the number of bins that has effects on accuracy and efficiency.} As the number of bins increases, within a certain range, the accuracy of GOOD increases. Note that, by increasing the number of bins, the computation time is also increased~\cite{Relate05}. The best results were achieved by setting the number of bins to $9$. The ESF descriptor does not have any parameter while VFH and GRSD have a \textit{radius} parameter which is used to estimate a normal vector for constructing a reference frame. We observed that setting the radius parameter to $0.006$ produced the best results. In the case of deep learning approaches, the best results ~{are obtained} by setting the resolution of the input image to $100 \times 100$ pixels. The $k$-parameter for the KNN classifiers was set to $3$ for all individual approaches, as we observed that there was generally no improvement in performance with larger values for $k$ for most of the approaches. The remaining parameters of all descriptors are defined as the default values. Results of the best configuration of each method with various distance functions are summarized in Fig.~\ref{acc_different_functions} (\textit{left} and \textit{center}). {A comparison of the results showed that most of the object representation approaches achieved a slightly better recognition accuracy based on the Motyka distance function}. Additionally, we measured the experiment time for each approach to define which approaches are computationally expensive (see Fig.~\ref{acc_different_functions} -- \textit{right}). Among handcrafted approaches, GRSD descriptor performed considerably worse in terms of recognition accuracy, while ESF, GOOD, and VFH representations demonstrated a good trade-off between recognition accuracy and experiment time. Particularly for real-time applications with limited resources, handcrafted approaches provide a favorable computation time. According to these experiments, Inception-ResNet and VGG$19$ were computationally expensive to use in ensemble learning methods for robotics applications. In contrast, ResNet$50$, Inception, and DenseNet offered an acceptable trade-off between recognition accuracy and experiment time (see Fig.~\ref{acc_different_functions} (\textit{center} and \textit{right})). {It can be
seen that out of all deep learning approaches, ResNet$50$ demonstrated the best computation time followed by Inception, and DensNet, respectively. We also observed that the computation time of deep learning approaches was significantly higher than handcrafted approaches.} From these results, we came up with three ensemble methods based on (\textit{i}) only handcrafted representations (ESF + GOOD + VFH), (\textit{ii}) only deep representations (ResNet$50$ + Inception + DenseNet), and (\textit{iii}) mixture of handcrafted and deep representations (ESF + GOOD + ResNet$50$). In the following subsections, we extensively evaluate all ensemble methods in different settings. 
%=====================================================

%=====================================================
\subsection {Scalability and Efficiency}
To evaluate the scalability of ensemble approaches, the synthetic RGB-D dataset is randomly divided into $10$, $30$, $50$, $70$ and $90$ categories. The performance of all methods ~{is} then measured based on ``\textit{average class accuracy}'' and ``\textit{computational time}'' (experiment time) achieved in a $10$-fold cross-validation procedure. In particular, we assessed the scalability as a function of object recognition accuracy versus the number of categories. Results are reported in Fig.~\ref{scalability_of_ensambles_plot}, Fig.~\ref{Time_Hc_Ensemble} and Table~\ref{scalability_of_ensambles}. By comparing all results, it is clear that the ensemble learning approaches outperformed individual methods in terms of object recognition accuracy. 

{Our results indicate that the ensemble methods consistently outperformed the individual methods, suggesting that the selected methods are complementary to each other rather than duplicating one another's performance.} On closer inspection, we can see that by introducing more categories, the accuracy of all approaches decreases. This is expected as the number of categories known by the system increases, ~{and the} classification task becomes more challenging. Among all approaches, DenseNet and ensemble with only deep representations always maintained accuracy above $90\%$ for all ~{levels} of scalability. Ensemble with deep-only representations had marginally higher accuracy than DenseNet. We also observed that the computation of all ensembles ~{is} substantially higher than individual base learners. Experimental results indicated that, among all ensemble methods, the handcrafted one achieved the best computation time followed by a mixture of handcrafted and deep representations. Moreover, we observed that the ensemble of deep-only representations is computationally very expensive. To form an ensemble of mixed representations, we considered GOOD, ESF, ~{and ResNet$50$} since these representations demonstrated a good balance between accuracy and computation time. Table~\ref{scalability_of_ensambles} indicates that the ensemble method based on ~{the} mixture of representations performed better than the ensemble with only handcrafted representations in all ~{levels} of scalability. According to these results (accuracy and computation time), ensemble learning based on multiple representations is a good choice for robotics applications, where resource budget becomes a valid consideration.

\begin{table*}[!t]
\newcolumntype{?}{!{\vrule width 0.5pt}}
\setlength\arrayrulewidth{0.5pt}
    \caption{Summary of scalability experiments~{on the synthetic household object dataset in offline scenario}.}
    \begin{center}
        \resizebox{0.9\linewidth}{!}{
        \begin{tabular}{|c|c|c|c|c|c|}
        \hline
        \multirow{2}{*}{\textbf{Ensemble Methods}}&\multicolumn{5}{c|}{\textbf{Accuracy with respect to varying numbers of
categories}} \\
        \cline{2-6} 
         & \textbf{{10 category}}& \textbf{{30 category}}& \textbf{{50 category}}& \textbf{{70 category}}& \textbf{{90 category}} \\
        \hline
        Ensemble (Handcrafted-only)&0.9318&0.9395&0.8694&0.8523&0.8228 \\
        \hline
        Ensemble (Deep-only)&  0.9717 & 0.9733 & 0.9155 & 0.9077 & 0.9177 \\
        \hline
        Ensemble (Mixed)& 0.9533& 0.9539& 0.8740& 0.8547& 0.8498 \\
        \hline
        \end{tabular}}
    \label{scalability_of_ensambles}
    \end{center}
\end{table*}

%=================================================
\subsection{Robustness}
\label{sec:robustness}

In real-world settings the test data differs from the training data because of several factors, e.g., sensor noise, domain shift, the gap between simulation and real-world data in case of sim-to-real transformation, and etc. The robustness of the ensemble and base learner approaches with respect to different levels of Gaussian noise~\cite{Result04} and varying point cloud resolutions~\cite{Relate05} was evaluated and compared. Similar to the previous round of experiments, we followed 10-fold cross validation and used restaurant object dataset for these tests.

\subsubsection{Gaussian noise}

\begin{figure}
    \begin{tabular}{cccc}
       \includegraphics[width=0.190\columnwidth]{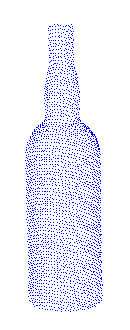}
       &
       \includegraphics[width=0.190\columnwidth]{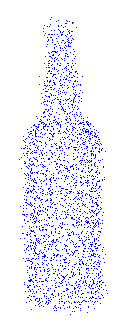}
       & 
       \includegraphics[width=0.190\columnwidth]{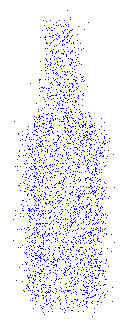}
       & 
       \includegraphics[width=0.190\columnwidth]{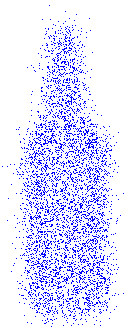}
       \\
       original & $\sigma=3$mm &$\sigma=6$mm &$\sigma=9$mm 
    \end{tabular}
	\caption{ An illustration of a bottle {from the synthetic household object dataset} with three different levels of Gaussian noise. We applied noise in all three axes of the object.}

\label{gaussian_object}
\end{figure}

To test the robustness against Gaussian noise, we performed $10$ rounds of evaluation, in which the ten levels of Gaussian noise with standard deviation ranging from $1$ mm to $10$ mm were added to the test data. In these experiments, Gaussian noise was independently added to each of the X, Y, and Z axes of the test object. To make the process clearer, we visualize a bottle object with three levels of Gaussian noise (i.e., $\sigma=3$mm, $\sigma=6$mm, $\sigma=9$mm) in Fig.~\ref{gaussian_object}. 

\begin{figure*}[!t]
    \begin{tabular}{ccc}
        \hspace{-0.35cm} \includegraphics[width=0.35\linewidth]{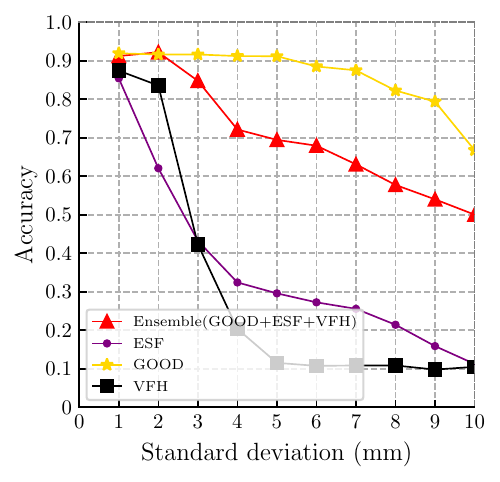} & \hspace{-0.6cm}
        \includegraphics[width=0.33\linewidth, trim= 0mm 0mm 0mm 0mm, clip = true]{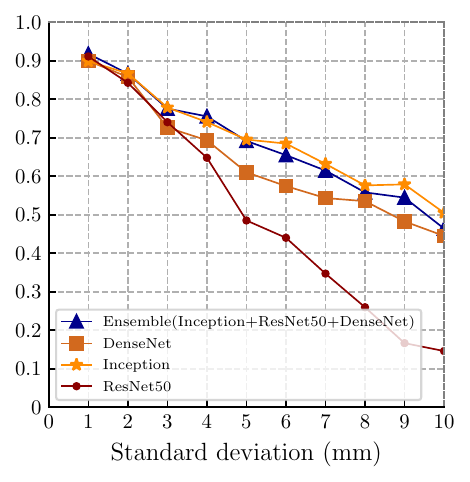} & \hspace{-0.6cm}
         \includegraphics[width=0.33\linewidth, trim= 0mm 0mm 0mm 0mm, clip = true]{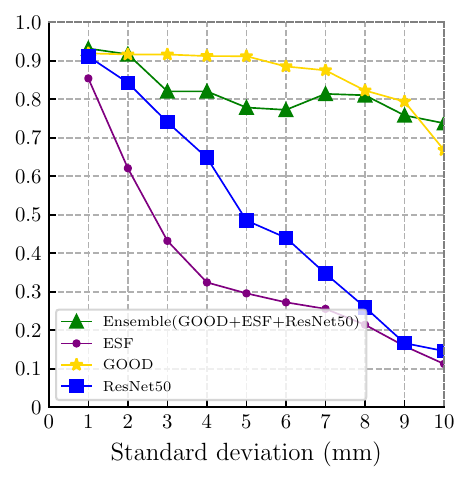}
    \end{tabular}
    \caption{Robustness of base learners and ensemble learning methods over different levels of Gaussian noise ~{ on the synthetic household object dataset in an offline scenario}: (\textit{left}) handcrafted approaches; (\textit{center}) deep learning methods; (\textit{right}) ensemble of mixture of handcrafted and deep learning approaches, and base learners.}
    \label{acc_gaussian}
\end{figure*}

The obtained results are plotted in Fig.~\ref{acc_gaussian}. One important observation is that the GOOD base learner showed a stable performance under all levels of noise, and outperformed all ensembles and base learners by a large margin. In contrast, the performance of all other base learners decreased drastically as the level of noise increased. Ensemble methods achieved the second best performance, while ResNet, ESF, and VFH were very sensitive to Gaussian noise. By comparing the results, it is visible that ensemble methods with handcrafted, and mixed of handcrafted and deep representations, showed exceptionally better accuracy than all base learners excluding the GOOD learner. The underlying reason was that a tie vote occurred frequently, and in most of the cases, the GOOD learner (a member of ensemble) won since it produced the closest representation to the query object. 

Such results are explained by the fact that GOOD descriptor uses a stable, unique and unambiguous object reference frame that is not affected by noise. We also observed that VFH is highly sensitive to noise as it relies on surface normal estimation to produce the representation of the object. Since ESF uses distances and angles between randomly sampled points to produce a shape description, it is sensitive to noise. In the case of ResNet$50$, by applying Gaussian noise to the point cloud of the object, the extracted images of the object are changed. Therefore, the deep representation of the object differs significantly from the original representation of the object, resulting in misclassification and poor performance. Overall, these results showed that ensemble learning using mixture of deep and handcrafted representations could yield better results.

%%%%%%%%%%%%  Wine Bottle Object _ Gaussian
\begin{figure}[!b]
\begin{tabular}{cccc}
   \includegraphics[width=0.20\columnwidth]{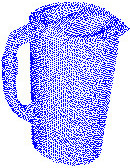}
   &
   \includegraphics[width=0.20\columnwidth]{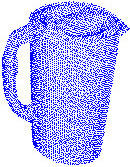}
   & 
   \includegraphics[width=0.20\columnwidth]{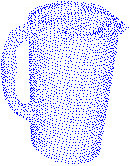}
   & 
   \includegraphics[width=0.20\columnwidth]{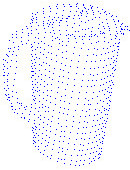}
   \\
    original & DS=1mm & DS=5mm & DS=10mm 
\end{tabular}
	\caption{An illustration of a pitcher~{from the synthetic household object dataset} with three different levels of downsampling.}

\label{down_sample}
\end{figure}
\begin{figure*}[!t]
    \begin{tabular}{ccc}
        \hspace{-0.35cm} \includegraphics[width=0.35\linewidth]{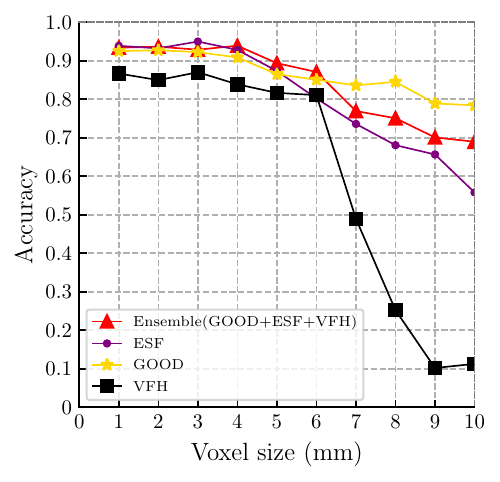} & \hspace{-0.6cm}
        \includegraphics[width=0.33\linewidth, trim= 0mm 0mm 0mm 0mm, clip = true]{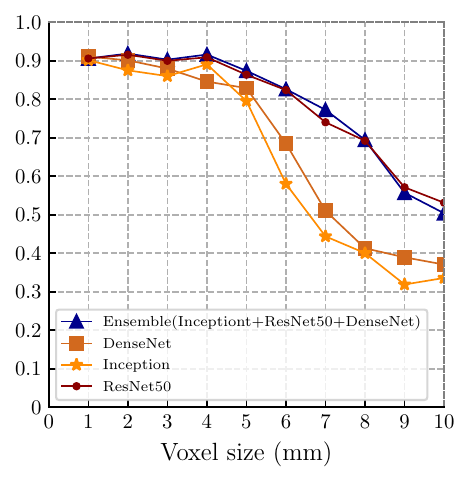} & \hspace{-0.6cm}
        \includegraphics[width=0.33\linewidth, trim= 0mm 0mm 0mm 0mm, clip = true]{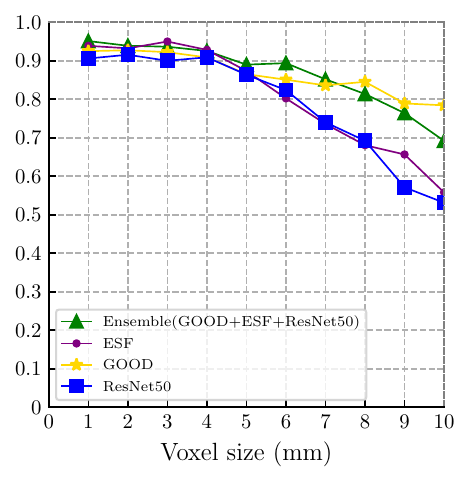}
    \end{tabular}
    \caption{Robustness of base learners and ensemble learning methods over various levels of downsampling~{ on the synthetic household object dataset in an offline scenario}: (\textit{left}) handcrafted approaches; (\textit{center}) deep learning methods; (\textit{right}) ensemble of mixture of handcrafted and deep learning approaches, and base learners.}
    \label{acc_downsample}
\end{figure*}

%################## downsampling
\subsubsection{Varying point cloud density}

To evaluate the robustness against varying point cloud density, two sets of experiments were carried out on the Restaurant object dataset. In these experiments, the density of train data was preserved as original while the density of test data was downsampled using ten different voxel sizes, ranging from $1$mm to $10$mm with a $1$mm interval. An illustration example of a Pitcher object with three levels of downsampling is shown in Fig.~\ref{down_sample}. 
We summarized the obtained results in Fig.~\ref{acc_downsample}. A comparison of all the results reveals that by increasing the downsampling resolution, the performance of all methods decreased significantly. Furthermore, we can see that all ensemble methods performed better than the individual base-learners ~{in low- and middle-resolution downsampling} (i.e., DS $\geq
6$mm), while in the case of high-resolution downsampling (i.e., DS $>
6$mm), the GOOD descriptor and ResNet$50$ showed slightly better performance than the ensemble learning methods.  Among handcrafted-based learners, GOOD  showed stable performance for all levels of downsampling while the performance of base learners with VFH and ESF descriptors dropped significantly under high levels of downsampling (see Fig.~\ref{acc_downsample} -- \textit{left}). It can be concluded from this observation that the GOOD descriptor is robust to downsampling mainly due to using a unique and stable reference frame, and normalized orthographic projection, while VFH relies on surface normals estimation and ESF computes several statistical features (i.e., distances and angles between randomly sampled points) to generate a shape description for a given point cloud. In the case of deep representations, all base learners performed very good under the low and mid level of downsampling. As the downsampling resolution was increased, the performance of all deep learning-based methods decreased (see Fig.~\ref{acc_downsample} -- \textit{center}). We also observed that the ensemble learning based on the mixture of handcrafted and deep representations (ResNet$50$ + ESF + GOOD) outperformed other ensemble methods under all levels of downsampling resolutions (see Fig.~\ref{acc_downsample} -- \textit{right}). 

\subsection{Evaluation Over Different Datasets}

\begin{figure}[!t]
    \begin{tabular}{cc}
        \hspace{-0.45cm} \includegraphics[width=0.53\columnwidth]{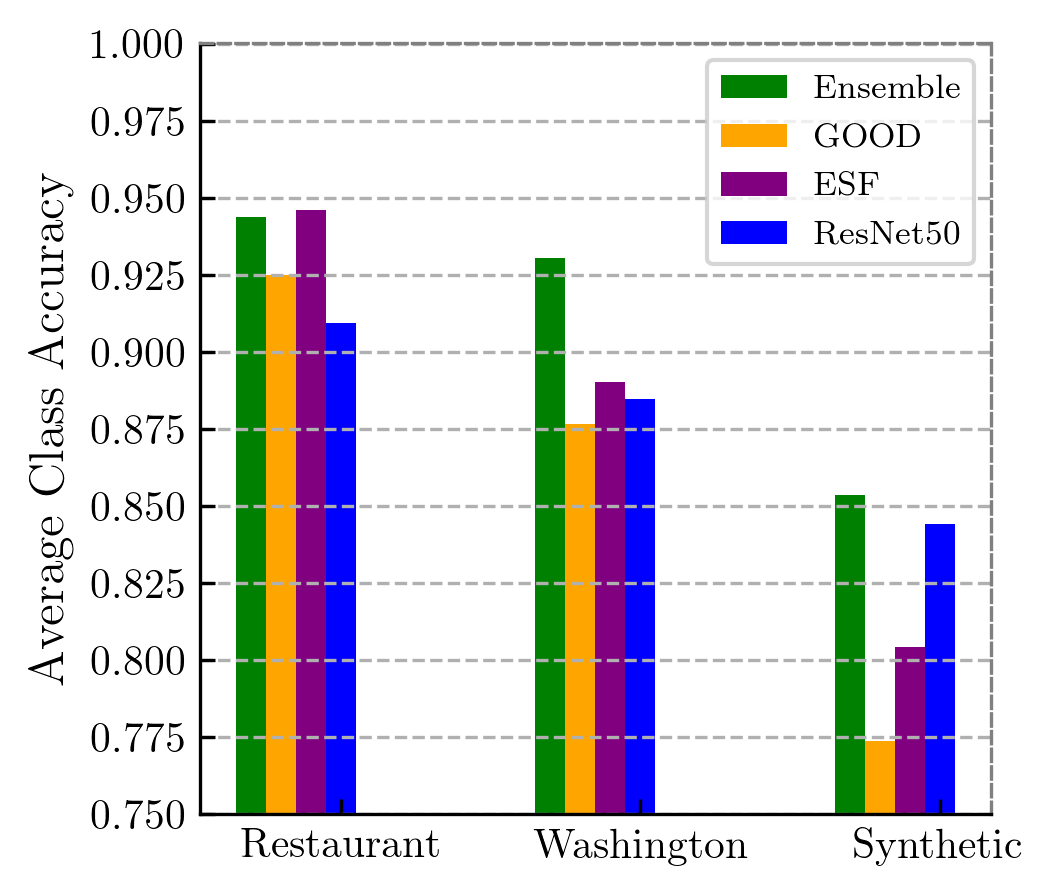} & \hspace{-0.8cm}
        \includegraphics[width=0.53\columnwidth, trim= 0mm 0mm 0mm 0mm, clip = true]{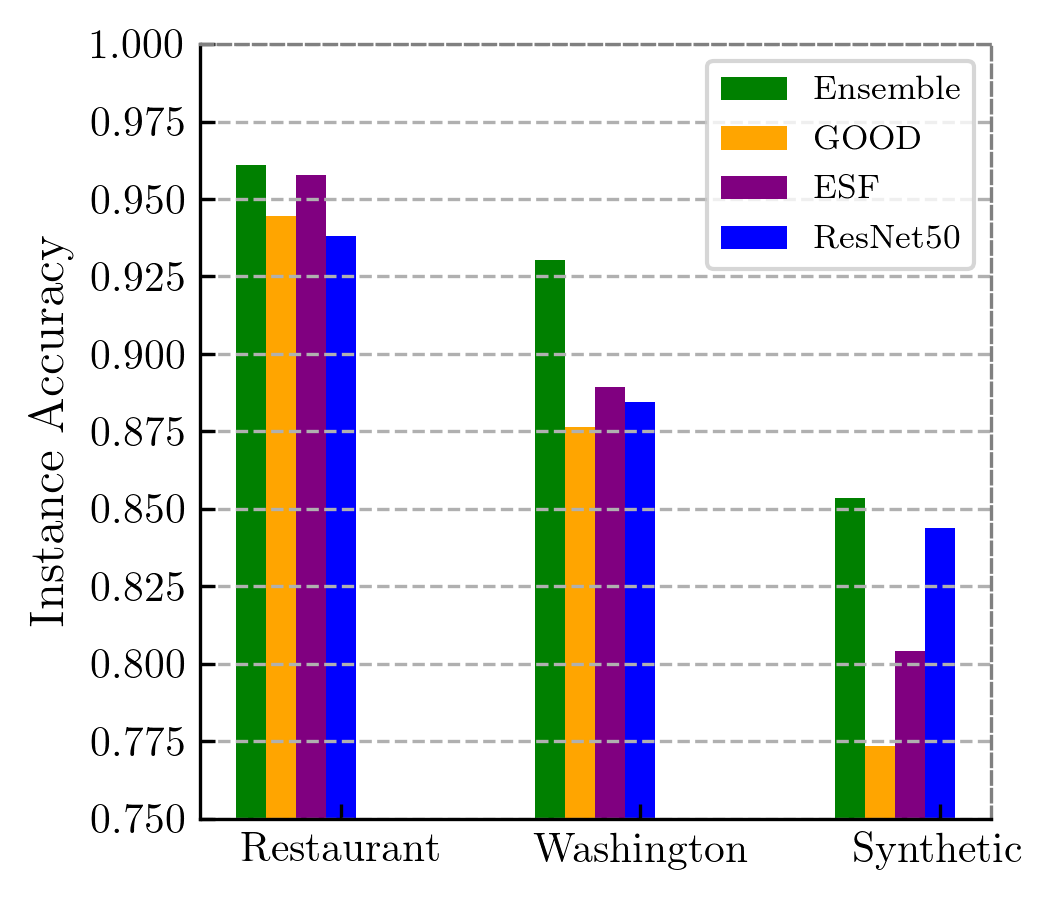} 
    \end{tabular}
    \caption{Object recognition performance of different approaches over various 3D object datasets {in offline scenarios}: (\textit{left}) average class accuracy; (\textit{right}) instance accuracy. Note that the ensemble method is based on a mixture of deep and handcrafted representations.}
    \label{different_datasets}
    \vspace{-4mm}
\end{figure}

We examined the performance of all approaches over three different 3D object datasets, including the synthetic dataset, Washington RGB-D object dataset~\cite{Result02} and Restaurant object dataset~\cite{Result07}. Since instance accuracy is susceptible to class imbalance, we report both instance and average class accuracies. ~{The results} are reported in Fig.~\ref{different_datasets}. In all experiments, ~{the ensemble method outperformed all base learners}. On closer inspection, we realized that on the Washington RGB-D dataset, ~{both instance and average class accuracies of the ensemble method were about $4\%$ higher than the best approach among GOOD, ESF, and ResNet$50$.} Even though ensemble and ESF methods achieved similar average class accuracy on the Restaurant object dataset (i.e. $94 \%$), the ensemble method showed a higher instance accuracy on the same dataset. Regarding the synthetic household object dataset, ensemble approaches achieved the best results, followed by ResNet$50$, ESF, and GOOD descriptors. Furthermore, since several object categories are geometrically very similar, base learners using shape-only descriptors (ESF and GOOD) could not achieve good accuracy on the Synthetic household object dataset.

%==============================================
\subsection{Open-ended Evaluation}

{In this round of experiments, we assessed the proposed approaches using an open-ended evaluation protocol~\cite{Result07,Result17,INTR03, Result18}. The main idea is to imitate the interaction of a robotic agent with the surroundings over a long period of time. 
In this open-ended scenario, the ensemble classifier is designed to adapt dynamically. Updates to the classifier occur either when a new category is introduced or in response to a misclassification. This flexible approach allows the system to continuously evolve and refine its classification capabilities, ensuring it remains responsive to changes in the dataset or instances where the current model performance falls short.} 

In particular, a \texttt{simulated user} introduces new object categories on demand using three randomly selected object views, in the same streaming fashion as would happen in any open-ended environment. The \texttt{simulated user} will then ask the agent to classify a set of never-before-seen instances of the known categories to determine whether the agent is accurate or not. The \texttt{simulated user} provides corrective feedback when the agent misclassifies an object. This way, the agent can improve the model of that specific category using the misclassified object (\texttt{{test-then-train}}).
The \texttt{simulated user} keeps training and evaluating the known category models until a certain protocol threshold $\tau$ has been reached, after which a new category is introduced. {We set $\tau$ to $0.67$, meaning that the object recognition accuracy of the agent is at least twice as high as its error rate. 
We discussed the effect of different protocol thresholds (stricter teacher) on agent performance (see subsection~\ref{protocol_threshold}).} 

\begin{figure}[!t]
\includegraphics[width=\linewidth]{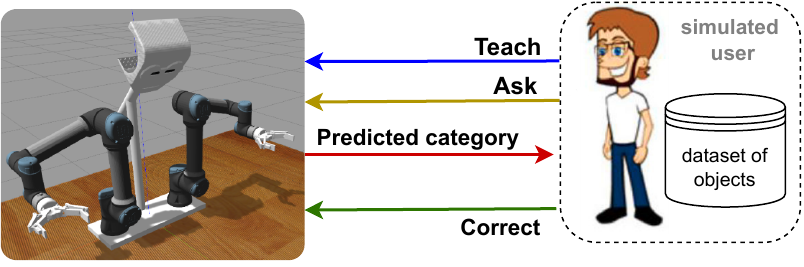}
\caption{Abstract architecture of interaction between the simulated user and the learning agent: The simulated user is connected to a large 3D object dataset and can interact with the learning agent using three actions: \textit{teach} to introduce ~{a new category}, \textit{ask} to assess the performance of the agent, and \textit{correct} to provide feedback in case of misclassification.}
\label{overview_sim_user}
\end{figure}

As soon as a new category is taught, the simulated user tests the agent's performance on all previously learned categories to ensure that no catastrophic interference has occurred. This way the agent could gradually increase its knowledge within a specific context. In case the model is unable to reach the protocol threshold after a certain number of question/correction iterations (QCI), e.g. $100$, \texttt{{simulated user}} stops the experiment, as the agent is unable to learn more categories. An alternative stopping condition would be the ``\textit{lack of data}'', where the agent learns all categories before reaching the point where no more categories can be learned (see Fig.~\ref{overview_sim_user}). Although a \texttt{{human user}} could also follow this protocol, following the protocol by a \texttt{{simulated user}} allows us to perform consistent and reproducible experiments in a fraction of the time a human would take to do the same experiment.

\subsubsection{Dataset and evaluation metrics} 

We used the synthetic household object dataset ($90$ categories), and the Washington RGB-D object dataset~\cite{Result02} ($51$ categories). We assess the performance of the agent using five metrics~\cite{Result22}:
\begin{itemize}
    \item Average number of question/correction iterations (QCI) required to learn the categories. {
    QCI measures the speed at which the agent learns a set of categories in an open-ended fashion, and a lower value indicates better performance. To ensure a fair comparison between different approaches, it is important to consider the number of categories learned, as learning a larger number of categories requires more QCI.}
    
    \item Average number of stored instances per category (AIC). Both QCI and AIC metrics indicate the time and memory usage required to learn ALC categories (the lower the better).
    \item Average number of learned categories (ALC) at the end of the experiment. {This metric is an indicator of how much the system was capable of learning (the higher the better).}
    \item Global classification accuracy (GCA), which describes the accuracy of a model over the whole run {(the higher the better)}.
    \item Average protocol accuracy (APA). The GCA and APA are ~{indicators} of how well the agent learns {(the higher the better)}.
\end{itemize}
To make a fair comparison, the order in which new instances and categories are introduced should be the same for all methods. In each round of experiments, the agent begins with no prior knowledge, and the order of introducing the instances and categories is randomly generated. The experiment is repeated ten times, and the average value of each metric is reported. 

\begin{figure*}[!t]
\includegraphics[width=\linewidth, trim= 70mm 0mm 60mm 0mm, clip=true]{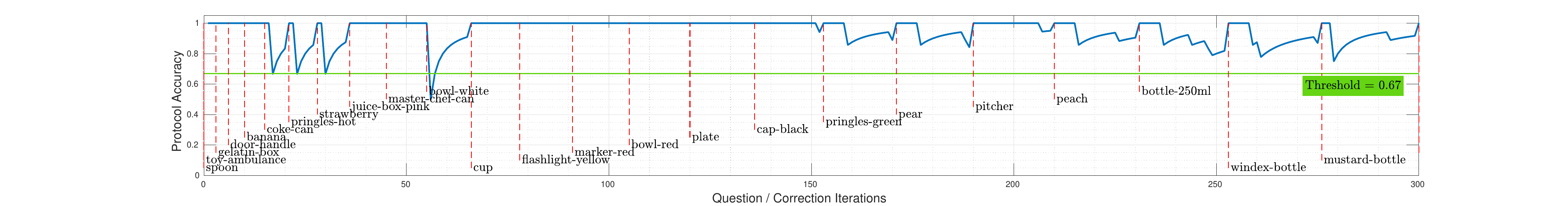}
\vspace{-5mm}
\caption{Evolution of protocol accuracy over first $300$ iterations~{on household objects in online scene}: in this experiment, the \texttt{\small{simulated user}} interactively teaches a new object category to the agent and evaluates its performance on all previously ~{learned categories}. In particular, if the performance of the agent exceeds a protocol threshold (i.e., set to $0.67$, as shown by the green horizontal dashed line), the \texttt{\small{simulated user}} introduces a new category (shown by a red dashed line and a category label) and examines the recognition accuracy of the agent on all previously learned categories using never-seen-before instances.}
\label{open_ended_process}
\end{figure*}

\subsubsection{Results}  To gain a better understanding of the open-ended learning process, we plotted the performance of the agent in the initial $300$ question/correction iterations in Fig.~\ref{open_ended_process}. Initially, the \texttt{\small{simulated user}} taught two object categories (\textit{spoon} and \textit{toy-ambulance}) to the agent, and progressively evaluated the classification accuracy of the agent after introducing a new object category. If the accuracy of the agent is greater than the protocol threshold (marked by the horizontal green line), a new category is introduced (\textit{gelatin-box}). The accuracy of the agent remained $100\%$ for the first five categories. After teaching the \textit{coke-can} category, some misclassification happened, and the \texttt{\small{simulated user}} provided corrective feedback accordingly. The agent then improved the model of the category and consequently improved its performance. This pattern repeated for three consecutive categories. Upon introducing the \textit{juice-box} and \textit{master-chef-can} categories, the agent recognized all known categories without any misclassification, whereas after teaching the \textit{bowl} category, some misclassifications occurred. Again, the agent improved its knowledge after receiving some feedback. Furthermore, it can be seen that this model could learn many more categories after the first $300$ question/correction iterations, given that the protocol accuracy consistently stays above the protocol threshold $\tau$. We observed that the agent could learn all $90$ synthetic household object categories after around $4120$ QCI iterations.

\begin{figure}[!t]
  \begin{tabular}{cc}
  \hspace{-4mm}
  \includegraphics[width=0.525\linewidth, trim= 0mm 0mm 8mm 5mm, clip = true]{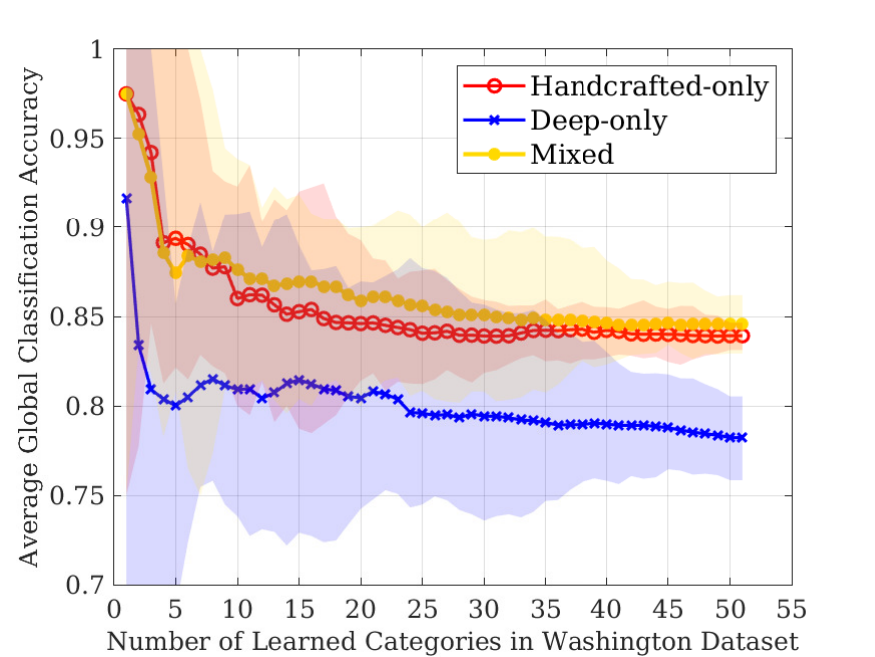} & \hspace{-6mm}
   \includegraphics[width=0.510\linewidth, trim= 5mm 0mm 8mm 5mm, clip = true]{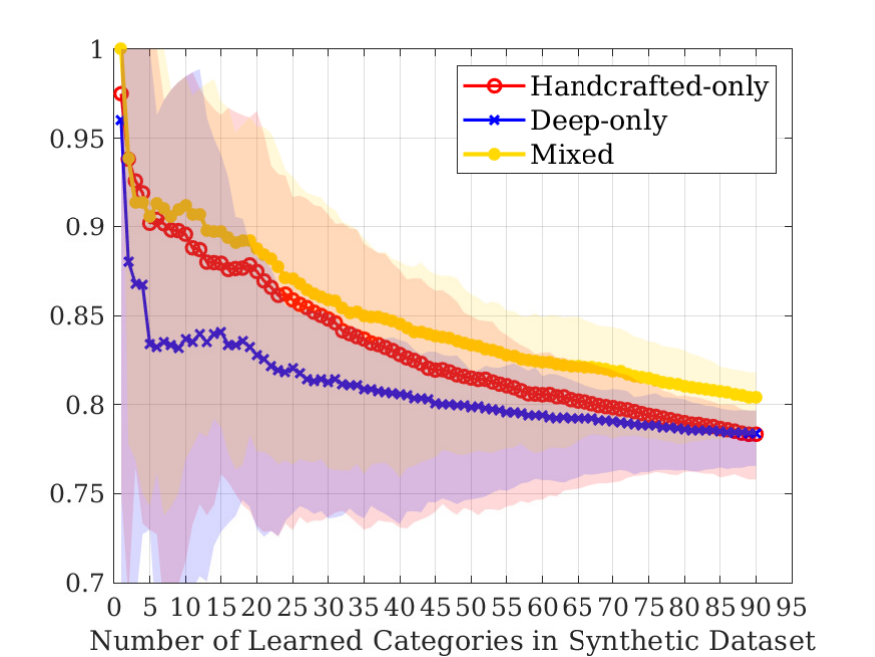}
    \end{tabular}
    \caption{The global classification accuracy as a function of the number of learn categories indicates how well the system learns in an open-ended fashion~{on the Washington and synthetic household objects datasets}. The performance of the learning agent over $10$ experiments is shown by the average (solid lines) and standard deviation (highlighted areas) on  (\textit{left}) Washington RGB-D object dataset; and (\textit{right}) Synthetic household object dataset. }
    \label{open-ended_evaluation_acc}
\end{figure}

\begin{table*}[!t]
   \newcolumntype{?}{!{\vrule width 0.5pt}}
\setlength\arrayrulewidth{0.5pt}
    \caption{Summary of open-ended evaluations {based on metrics described in Section 4.6.1}.}
    \begin{center}
        \resizebox{0.8\linewidth}{!}{
        \begin{tabular}{|c|c|c|c|c|c|c|}
        \hline
        \multirow{2}{*}{\textbf{Dataset}}&\multirow{2}{*}{\textbf{Methods}}&\multicolumn{5}{|c|}{\textbf{Metrics of open-ended evaluations}} \\
        \cline{3-7} 
         & &\textbf{QCI}$\mathbf{\downarrow}$ & \textbf{ALC}$\mathbf{\uparrow}$ & \textbf{AIC}$\mathbf{\downarrow}$ & \textbf{GCA}$\mathbf{\uparrow}$ & \textbf{APA}$\mathbf{\uparrow}$\\
        \hline  \hline

        \multirow{5}{*}{\textbf{Washington}}&RACE~\cite{Result20}&382.10&19.90&8.88&0.67&0.78 \\
        \cline{2-7} 
        & BoW~\cite{Result19}&  411.80 & 21.80 & 8.20 & 0.71 & 0.82 \\
        \cline{2-7} 
       & Open-Ended LDA~\cite{Result21}& \textbf{262.60}& 14.40& 9.14& 0.66& 0.80\\
       \cline{2-7} 
       & Local-LAD~\cite{Relate16}&613.00&28.50&9.08&0.71&0.80\\
       \cline{2-7} 
       & OrthographicNet~\cite{Result18}& 1342.60&\textbf{51.00}& 8.97& \textbf{0.77}& \textbf{0.80}\\
       \cline{2-7} 
       &  ~{BEiT}~\cite{2021beit}& 1641.00&45.8& 13.8& 0.69& 0.75\\
       \cline{2-7} 
       & ~{Swin}~\cite{liu2021swin}& 1683.00&45.70& 14.17& 0.69& 0.74\\
       \cline{2-7} 
       &  ~{ConvNet}~\cite{liu2022convnet}& 1629.00&45.10& 14.06& 0.68& 0.74\\

       \hline \hline
       \multirow{3}{*}{\textbf{Washington}}& Ensemble (deep-only)& 1346.30&\textbf{51.00}& 8.42& 0.79& 0.82\\
       \cline{2-7} 
       &Ensemble (handcrafted-only)& \textbf{1329.40}&\textbf{51.00}& 7.70&0.82&0.84\\
       \cline{2-7} 
       &Ensemble (mixed)& 1330.10&\textbf{51.00}& \textbf{7.26}&  \textbf{0.84}&  \textbf{0.86}\\
       \hline  \hline
       \multirow{3}{*}{\textbf{Synthetic}}& Ensemble (deep-only) & 4122.70&\textbf{90.00}& 12.90& 0.78& 0.80\\
       \cline{2-7} 
       &Ensemble (handcrafted-only)& 4114.00&\textbf{90.00}&12.90& 0.78&0.81\\
       \cline{2-7} 
       &Ensemble (mixed) & \textbf{4103.40}&\textbf{90.00}& \textbf{11.92}& \textbf{0.80} & \textbf{0.83}\\
       \hline 
       \end{tabular}}
    \label{all_open-ended_experiments}
    \end{center}
\end{table*}

\begin{figure*}[!t]
\includegraphics[width=\linewidth, trim= 0mm 0mm 0mm 0mm, clip=true]{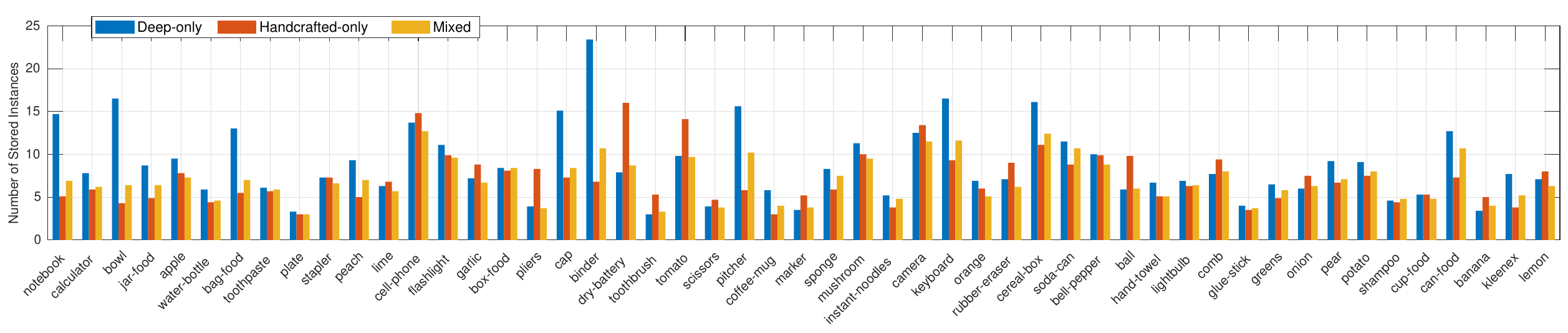}\\
\includegraphics[width=\linewidth, trim= 0mm 0mm 0mm 2mm, clip=true]{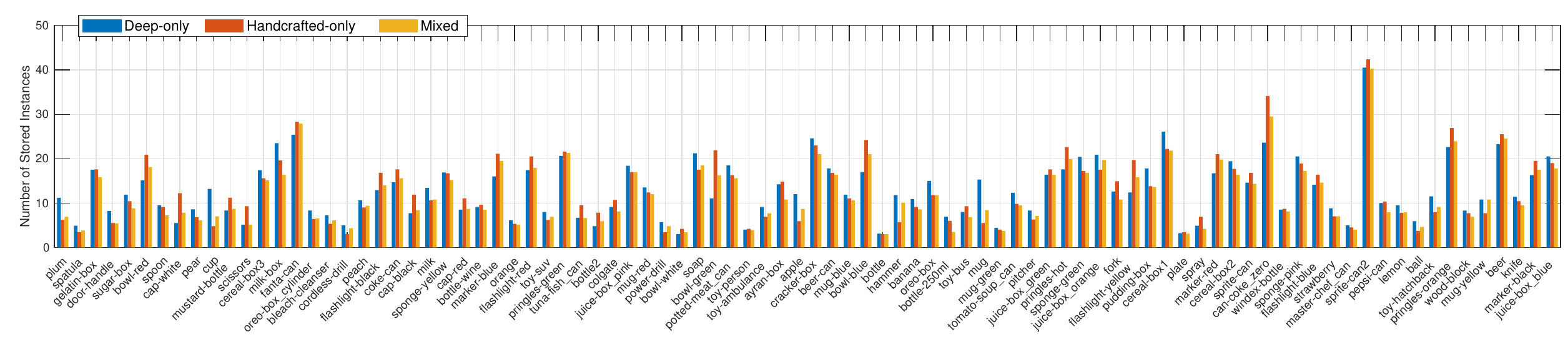}
\caption{ Average number of stored instances per category for ensemble learning based on handcrafted-only, deep-only, and a mixture of handcrafted and deep representations on two datasets~{in online scenarios}: (\textit{top-row}) experiments using the Washington RGB-D object dataset; (\textit{lower-row}) experiments using the synthetic household object dataset.}
\label{open_ended_stored_instances}
\end{figure*}

We performed $10$ round of experiments to evaluate the performance of the proposed ensemble approaches. A detailed summary of all experiments is reported in Table~\ref{all_open-ended_experiments}. Figure~\ref{open-ended_evaluation_acc} shows the recognition performance of the agent versus the number of categories learned. It can be concluded from this plot that the global accuracy of all approaches decreases as the number of categories ~{increases}. This is expected since the object recognition task is harder when there are more categories. We also observed that all approaches showed a good performance and the global classification accuracy of the agent remained above the protocol threshold. In addition, the ensemble approach with mixed representations achieved better global classification accuracy than the handcrafted-only and deep-only approaches.~{This experimental result also can be predicted. Deep neural network models encounter issues like catastrophic forgetting and overfitting~\cite{wang2023comprehensive,xie2021artificial}. Overfitting is particularly problematic when dealing with small datasets where the model tries to learn too many features. However, handcrafted descriptors possess a unique ability to extract specific features, especially in cases of limited data. When deep networks struggle to provide a descriptive representation for a given input {due to domain shift (i.e., natural images vs. orthographic images) or noisy orthographic images}, handcrafted descriptors can compensate for these shortcomings. As a result, combining handcrafted 3D descriptors with deep representations yields strong performance in open-ended learning scenarios. Additionally, the results in Table~\ref{all_open-ended_experiments} confirm this notion. The data clearly shows that ensemble learning demanded the largest average number of instances per category in order to effectively learn all classes within both the Washington dataset and the synthetic dataset.}

Figure~\ref{open_ended_stored_instances} shows the average number of instances stored in each category using (\textit{top-row}) Washington and (\textit{lower-row}) Synthetic household object datasets. In these plots, each bar represents the accumulation of three instances provided at the time the category was introduced and the instances that were corrected by the \texttt{\small{simulated user}} somewhere during the experiment. By comparing the results, it is visible that, on average, the ensemble with deep-only representations stored more instances than the ensemble with mixed and ensemble with hand-crafted-only representations (see Table~\ref{all_open-ended_experiments}). 

{ 
 As shown in Table~\ref{all_open-ended_experiments}, the number of iterations tends to increase as the number of learned objects increases, which implies a decrease in QCI performance. This phenomenon can be attributed to the increased complexity of the feature space with more learned objects, which makes learning new objects more difficult. Upon closer inspection of Table 3, it can be seen that the handcrafted-only approach outperforms the deep-only approach slightly in terms of QCI. This can be explained by the differences between their attributes. The ensemble approach of handcrafted-only encompasses both local and global information. In contrast, the ensemble deep-only approach predominantly emphasizes the locality, which accounts for the handcrafted-only better QCI performance on a small dataset. In addition, Figure~\ref{open_ended_stored_instances} also further reveals that during the early stages of learning, the deep-only approach requires more objects to be stored to learn a new instance compared to the handcrafted-only.}

It can be concluded that the ensemble ~{approaches} with hand-crafted-only and mixed representations outperform the ensemble with deep-only representations in terms of QCI (time) and AIC (memory) metrics. On a closer inspection, we observed that as the number of categories increased, misclassifications happened more frequently. Consequently, the agent received more corrective feedback to update the model of categories. It is worth mentioning that the Synthetic dataset contains fine-grained objects, such as a \textit{Pringles-onion} vs. \textit{Pringles-hot}. Such similarities undoubtedly make the classification task more challenging. Therefore, on average, experiments on the Synthetic dataset took longer QCI. 

We also compared the open-ended performance of our methods with five state-of-the-art approaches using the Washington dataset. Results are summarized in Table~\ref{all_open-ended_experiments}. We observed that RACE~\cite{Result20}, BoW~\cite{Result19}, Open-ended LDA~\cite{Result21}, Local-LDA~\cite{Relate16}, ~{Swin~\cite{liu2021swin}, BEiT~\cite{2021beit}, and ConvNet~\cite{liu2022convnet}} were not able to learn all categories. Furthermore, the obtained results indicated that all of our ensemble methods and OrthographicNet~\cite{Result18} were able to learn all $51$ categories and achieved a significantly higher GCA accuracy compared to the protocol threshed ($0.67$). According to these results, these methods are able to learn many more categories. Regarding QCI, the ensemble with mixed representations was able to learn all categories slightly faster than the other methods (i.e., required fewer iterations to learn all the categories). In terms of AIC, ensemble methods outperformed other approaches. More specifically the agent with the ensemble methods of handcrafted-only and mixed representations, on average, required $7.70$ and $7.26$ instances per category 
to learn all $51$ categories, while the agent with OrthographicNet and the ensemble of deep-only representations required $8.97$ and $8.88$ instances per category, respectively.

\subsubsection{Effect of protocol threshold on performance}
\label{protocol_threshold}
The protocol threshold ($\tau$) defines how accurately the agent should learn object categories. In particular, $\tau$ is the threshold accuracy which must be reached before a new category is
introduced by the \texttt{\small{simulated user}}. Therefore, we performed two ~{sets} of experiments by setting the $\tau \in \{0.8, 0.9\}$ to see how well the agent performed with a \texttt{\small{stricter simulated user}}. Results are reported in Table~\ref{different_thresholds_open-ended}. By setting the $\tau=0.8$, although our ensembles still could learn all $51$ categories, the ensemble of deep-only representations requires more iterations than the ensemble with mixed and handcrafted-only representations. 
\begin{table*}[!t]
\newcolumntype{?}{!{\vrule width 0.5pt}}
\setlength\arrayrulewidth{0.5pt}
    \caption{Summary of open-ended evaluations using two protocol thresholds on Washington RGB-D object dataset.}
    \vspace{-5mm}
    \begin{center}
        \resizebox{0.7\linewidth}{!}{
        \begin{tabular}{|c|c|c|c|c|c|c|}
        \hline
        \multirow{2}{*}{\textbf{Threshold}}&\multirow{2}{*}{\textbf{Method}}&\multicolumn{5}{|c|}{\textbf{Metrics of open-ended evaluations}} \\
        \cline{3-7} 
        & &\textbf{QCI}$\mathbf{\downarrow}$ & \textbf{ALC}$\mathbf{\uparrow}$ & \textbf{AIC}$\mathbf{\downarrow}$ & \textbf{GCA}$\mathbf{\uparrow}$ & \textbf{APA}$\mathbf{\uparrow}$\\
        \hline
        \hline
        \multirow{3}{*}{\textbf{0.80}} &Ensemble (deep-only)
        &1629.00 & \textbf{51.00} & 8.49 & 0.83 & 0.86\\
        \cline{2-7} &Ensemble (handcrafted-only) & 1491.00 & \textbf{51.00} & 7.53 & \textbf{0.85} &0.86 \\
        \cline{2-7} 
        &Ensemble (mixed)
        &\textbf{1415.00} & \textbf{51.00} & \textbf{7.14} & \textbf{0.85} & \textbf{0.87}\\
        \cline{2-7}
        
        \hline  \hline

        \multirow{3}{*}{\textbf{0.90}} 
        &Ensemble (deep-only) &1347.00&27.00&9.07&0.88&0.93\\\cline{2-7}
        &Ensemble (handcrafted-only) &\textbf{1335.00} &30.00 &8.00 &\textbf{0.89} &\textbf{0.94}\\
        \cline{2-7} 
        &Ensemble (mixed)
        &1633.00& \textbf{36.00}& \textbf{7.78}& \textbf{0.89} & \textbf{0.94}\\
        \hline
        \end{tabular}}
    \label{different_thresholds_open-ended}
    \end{center}
    \vspace{-5mm}
\end{table*}
These results indicate that the ensemble with mixed representations showed a better performance than ensemble methods with deep-only and handcrafted-only representations. Further increasing the protocol threshold to $0.9$ made learning new categories harder, and the agent could only learn $27$, $30$, and $36$ categories with deep-only, handcrafted-only, and mixed representations, respectively. 

\subsection{Robotic demonstrations}
\begin{figure}[!t]
    \centering
    \includegraphics[width=\linewidth]{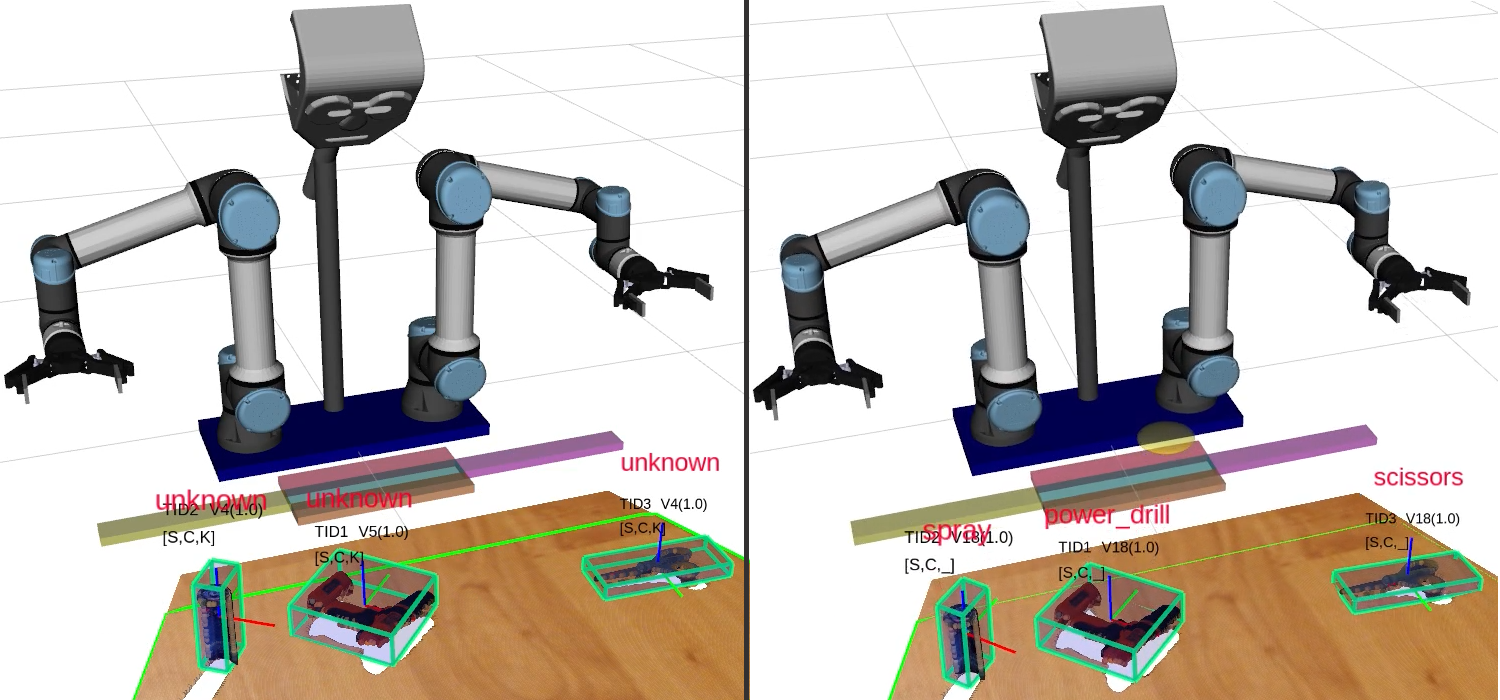}
    \vspace{-4mm}
    \caption{The recognition results of the robot during the ``\textit{set-the-table}'' task: (\textit{left}) initially the robot recognized all objects as \textit{unknown} since it did not have any information about the objects; (\textit{right}) the user then taught the objects to the robot using the graphical menu, and the robot conceptualized and recognized all object correctly.}
    \label{sim_exps_recognition}
\end{figure}
\begin{figure*}[!t]
\includegraphics[width=\linewidth, trim= 0mm 0mm 0mm 0mm, clip=true]{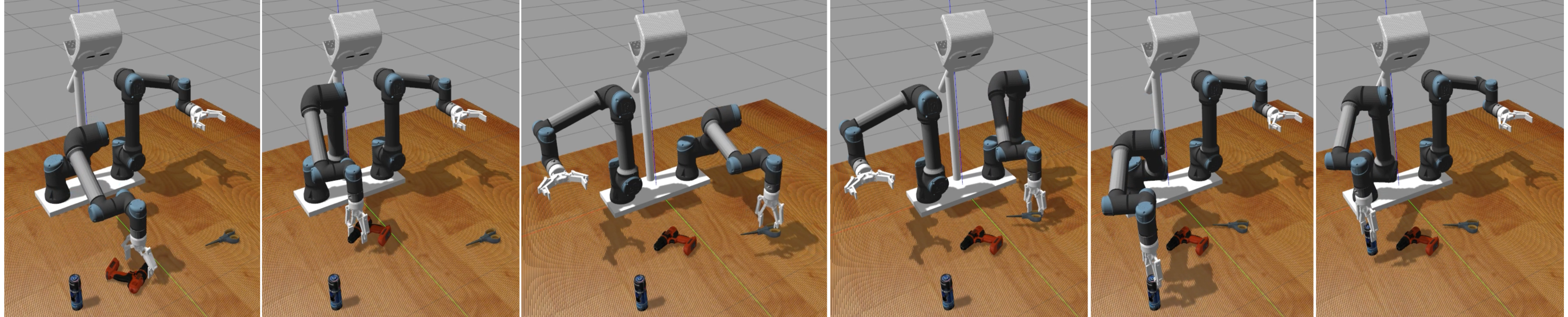}
\caption{A sequence of snapshots showing the performance of our dual-arm robot in \textit{``set-the-table''} task: Initially, the robot does not have any information about the objects. We randomly place three tool objects (e.g., \textit{spray}, \textit{power\_drill}, \textit{scissors}) on top of the table, and then teach the robot about the objects using a graphical user interface. The robot learns and recognizes all object categories correctly. Afterward, we instruct the robot to perform \textit{``set-the-table''} task. The robot then iteratively picks the nearest object to the robot and places it in a pre-defined position.}
\label{set_table_sim}
\end{figure*}

\begin{figure*}[!t]
\includegraphics[width=\linewidth]{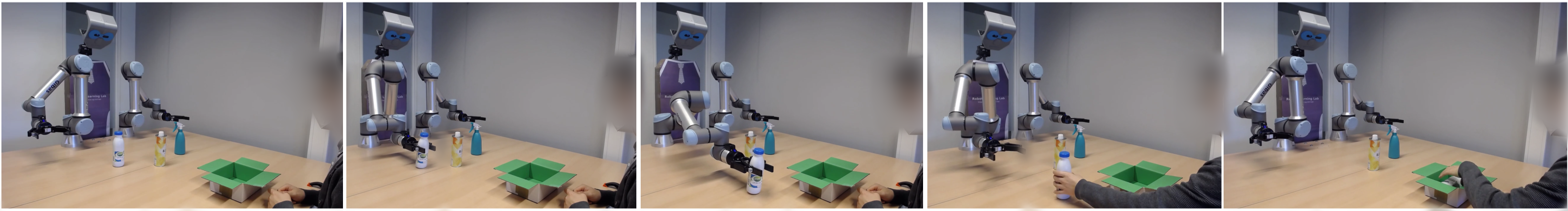}
\caption{A series of snapshots demonstrating the performance of our dual-arm robot in a robot-assisted packaging scenario: We used three objects that are geometrically similar (i.e., cylindrical) but have different textures. The objects are not reachable by human ~{users}, and therefore, the robot should ~{hand over} the requested object to the user. The robot must detect the pose and label of every object correctly to perform the experiment successfully.}
\label{real_robot_exps}
\end{figure*}

To show the real-time performance of the proposed ensemble of mixed representations and {the efficacy of our approach in concurrently handling multiple objects}, we integrate our approach into a robotic system~\cite{Result19}\cite{kasaei2021mvgrasp}. We performed two sets of robotic demonstrations, one set in simulation, and the other set on a real robot. Our simulated and real-robot experimental setups are shown in Fig.~\ref{set_table_sim} and Fig.~\ref{real_robot_exps}, respectively. In these experiments, the robot initially did not know any objects in advance and hence, recognized all objects as ``\textit{unknown}'' (see Fig.~\ref{sim_exps_recognition} -- \textit{left}). A human user then introduced the objects to the robot using a graphical menu. The robot conceptualized all objects and could recognize them correctly (see Fig.~\ref{sim_exps_recognition} -- \textit{right}). 
To complete a task successfully, the robot should grasp and manipulate the target object to the desired location.

In the case of simulation experiments, we considered the ``\textit{set-the-table}'' task. In particular, we randomly placed three objects (e.g., \textit{spray}, \textit{power\_drill}, \textit{scissors}) on top of the table and then instructed the robot to perform ``\textit{set-the-table}'' task by putting all objects in predefined locations (see Fig.~\ref{set_table_sim}). We repeated this experiment $20$ times with diffident sets of objects. We observed that the robot was able to learn and recognize all objects precisely and place them in the desired locations.  

\begin{figure}[!t]
    \centering
    \includegraphics[width=\linewidth]{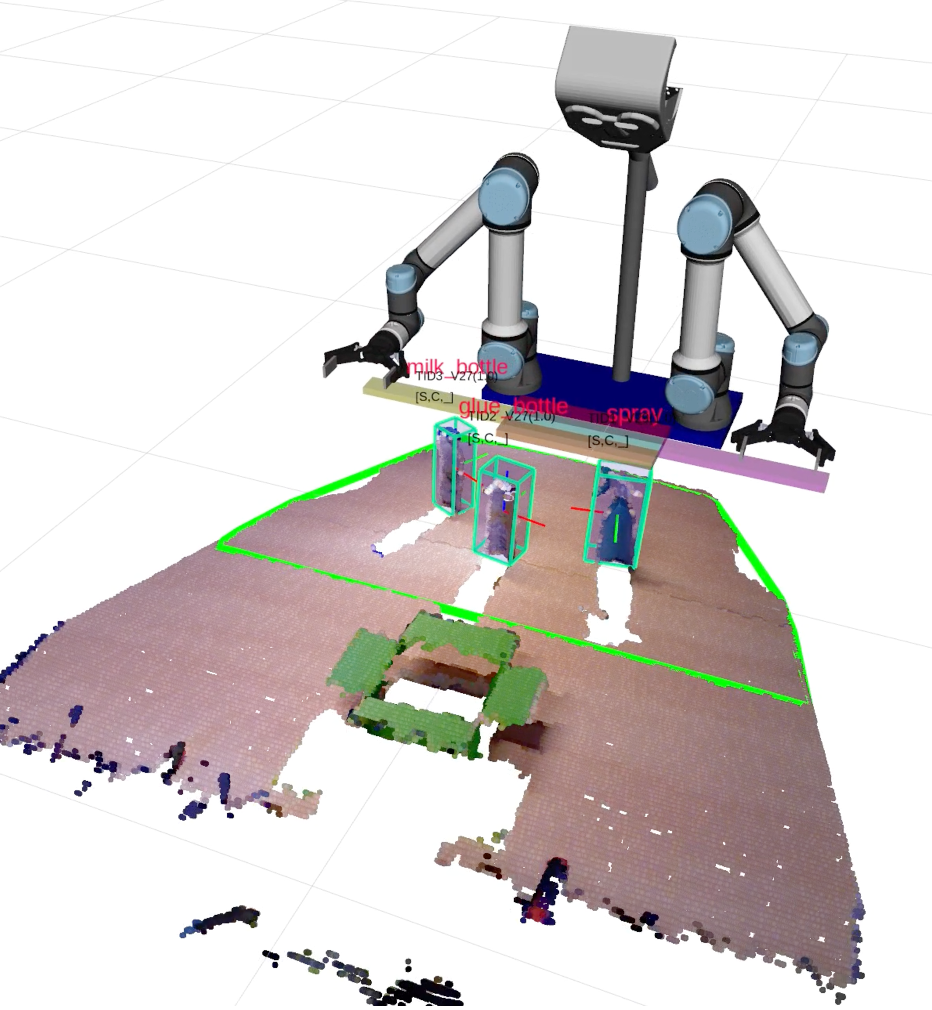}
    \caption{The perception of the robot during the ``\textit{robot-assisted-packaging}'' scenario: The robot's workspace is shown by the green convex hull. The pose of each object is shown by the bounding box and its reference frame. The recognition results are visualized above each object.  }
    \label{real_robot_perception}
\end{figure}

In real-robot experiments, we evaluated the proposed approach in the context of ``\textit{robot-assisted-packaging}'' scenarios, where our dual-arm robot ~{hands over} a target object to a human co-worker for packaging purposes (see Fig.~\ref{real_robot_exps}). In this round of experiments, we randomly placed three objects on top of the table which are initially not reachable by the human user. The selected objects were geometrically very similar (e.g., \textit{spray}, \textit{milk\_bottle}, and \textit{glue\_bottle}), making the few-shot learning and classification task more challenging (see Fig.~\ref{real_robot_perception}). In these experiments, the user first taught the label of objects to the robot, and then iteratively asked the robot to hand over a specific object. To accomplish the task successfully, the robot must initially detect and recognize all objects correctly, then pick the target object, and deliver it to the user. The user finally put the object into the box. Note that the robot used its right arm to manipulate the target object if it was on the right side of the robot, otherwise, the robot used its left arm. We repeated this experiment with various sets of objects $10$ times. We observed that the robot was able to learn and recognize all objects correctly and deliver them to the user successfully. 

{Throughout our experiments, we deliberately stopped the object tracker before executing the grasp action. This step is crucial, as otherwise tracker will be confused since the gripper, the robot's arm, and the object will be connected to each other after grasping and forming a large object. It is possible that the tracker fails if the object is entirely occluded by the robot for a brief period. However, once the object becomes visible again, the tracker will be able to detect it, and the robot will accurately recognize it.} A video of these experiments is available online at: \href{https://youtu.be/nxVrQCuYGdI}{\texttt{\small{https://youtu.be/nxVrQCuYGdI}}}

\section{Conclusions }

In this paper, we present lifelong ensemble learning approaches based on multiple representations to handle few-shot object recognition. More specifically, we formulate several ensemble methods based on deep-only, handcrafted-only and a mixture of handcrafted and deep representations. To facilitate open-ended learning, each of the base learners has a memory unit to store and retrieve object category information instantly. To assess the proposed approach in offline, and open-ended settings, we generated a large synthetic object dataset, consisting of $27000$ views of $90$ household object categories. Furthermore, we used two real 3D object datasets. Experimental results showed the effectiveness of the ensemble approaches on (few-shot) 3D object recognition tasks, as well as their superior lifelong learning performance over the state-of-the-art approaches. Among the ensemble learning methods, although the ensemble method with deep-only representations achieved slightly better recognition accuracy, it was computationally very expensive for robotics applications~\cite{ganaie2022ensemble}. We observed that the ensemble method with a mixture of handcrafted and deep representations showed a better trade-off between accuracy and computation time. We also demonstrated the real-time performance of the proposed approach in a set of real and simulated robotics experiments. In the simulation, we assessed the performance of the robot in the context of the ``\textit{set-the-table}'' task, while in the case of real-robot, we considered a ``\textit{robot-assisted-packaging}'' scenario. In both sets of experiments, the robot could learn and recognize all objects correctly and accomplish the tasks successfully. 

In the continuation of this work, we would like to investigate the possibility of improving performance by considering contextual information in ensemble methods.
Visual grounding and reasoning would be another interesting avenue for future research, where the robot segments an object from a crowded scene given a natural language description.

\section*{Acknowledgment}

We thank the Center for Information Technology of the University of Groningen for their support and for providing access to the Peregrine high-performance computing cluster. Songsong Xiong is funded by the China Scholarship Council.

\bibliographystyle{cas-model2-names}
\bibliography{reference}

\end{document}